\UseRawInputEncoding
\documentclass[journal]{IEEEtran}
\usepackage{graphicx}
\usepackage{amsmath,amssymb}
\usepackage{cases}
\usepackage{textcomp}
\usepackage{xcolor}
\def\BibTeX{{\rm B\kern-.05em{\sc i\kern-.025em b}\kern-.08em
    T\kern-.1667em\lower.7ex\hbox{E}\kern-.125emX}}
\usepackage{balance} 
\usepackage{bbm}
\usepackage{amsfonts}
\usepackage{epstopdf}
\usepackage{array}
\usepackage{booktabs}
\usepackage{color}
\usepackage{xcolor}
\usepackage{colortbl}
\usepackage{xspace}
\usepackage{multirow}
\usepackage{pifont}
\usepackage{bbding}
\usepackage{fontawesome}
\usepackage{hyperref}
\usepackage{caption}
\DeclareCaptionLabelSeparator{period}{. }
\usepackage[final,nopatch=footnote]{microtype}
\usepackage{makecell}
\usepackage{microtype}
\usepackage{caption,setspace}
\usepackage{hyperref}
\usepackage{algorithm}
\usepackage{algorithmic}
\usepackage[normalem]{ulem}

\AtBeginDocument{%
  \providecommand\BibTeX{{%
    Bib\TeX}}}
\begin{document}

\title{Towards Unbiased Federated Graph Learning: Label and Topology Perspectives}

\author{
Zhengyu Wu, 
Boyang Pang, 
Xunkai Li, 
Yinlin Zhu, 
Daohan Su, 
Bowen Fan, \\
Rong-Hua Li, 
Guoren Wang, 
Chenghu Zhou
        % <-this % stops a space


}

\maketitle

\begin{abstract}
    As a privacy-preserving collaborative framework, Federated Graph Learning (FGL) facilitates distributed training of graph neural networks without sharing data. 
    In our investigation, subgraph-FL has emerged as the dominant paradigm in FGL, with most recent efforts concentrating on enhancing overall node classification performance.
    However, due to the inherent complexity of node profiles (i.e., features and labels) and graph topology, these methods often overlook fairness considerations.
    Specifically, they exhibit biased performance to nodes with disadvantageous properties, such as being minority-class within local subgraphs or heterophilous connections (i.e., neighboring nodes possess dissimilar labels and misleading features).
    This underexplored fairness challenge reveals the robust concerns of current subgraph-FL methods: high accuracy conceals degraded performance on structurally or semantically marginalized node groups. 
    To address this, we advocate for
    (1) enhancing the representation of minority-class nodes for class-wise fairness and
    (2) mitigating topological biases arising from heterophilous connections for topology-aware fairness.
    In this context, we propose FairFGL, a novel framework that performs fine-grained mining of graph properties and orchestrates a collaborative learning paradigm to enhance fairness. 
    Specifically, on the client side, the History-Preserving Module prevents local models from overfitting to locally dominant classes.
    The Majority Alignment module mitigates topological bias by refining heterophilous majority-class representations.
    The Gradient Modification Module improves class-wise fairness by transferring minority-class knowledge from structurally favorable clients.
    On the server side, FairFGL uploads only the most influenced subset of locally trained parameters that best capture local data distribution while optimizing communication efficiency.
    Based on this, a cluster-based aggregation strategy reconciles conflicting updates and suppresses global majority dominance.
    Extensive evaluations on eight benchmark datasets show that FairFGL significantly improves performance for disadvantaged node groups—achieving up to a 22.62\% increase in Macro-F1—while enhancing convergence efficiency over SOTA baselines.
\end{abstract}

\section{Introduction}
    Graphs excel at modeling complex interactions among real-world entities and have demonstrated their effectiveness and versatility across diverse domains~\cite{infectious_Disease_Dynamics, xia2023app_gnn_rec1, binwu_fina1, hyun2023app_gnn_fina2}. 
    Graph Neural Networks (GNNs) have proven to be powerful for generating effective node embeddings by aggregating information from neighboring nodes, enabling remarkable performance~\cite{Hu2021ahgae, wu2019sgc,cai2021link_prediction2, tan2023link_prediction4, zhang2019graph_classification1,yang2022graph_classification3}. 
    To comply with both increasing privacy concerns against raw data acquisition and untransparency in growing business competition, Federated Graph Learning (FGL) emerged as a burgeoning field whose distributed settings of data storage and collaborative training architecture strike optimal balance in maintaining performance utility and data security. 

\begin{figure}[t]
\centering
    \setlength{\abovecaptionskip}{0.2cm}
    \setlength{\belowcaptionskip}{-0.5cm}
  \includegraphics[width=\linewidth,scale=1.0]{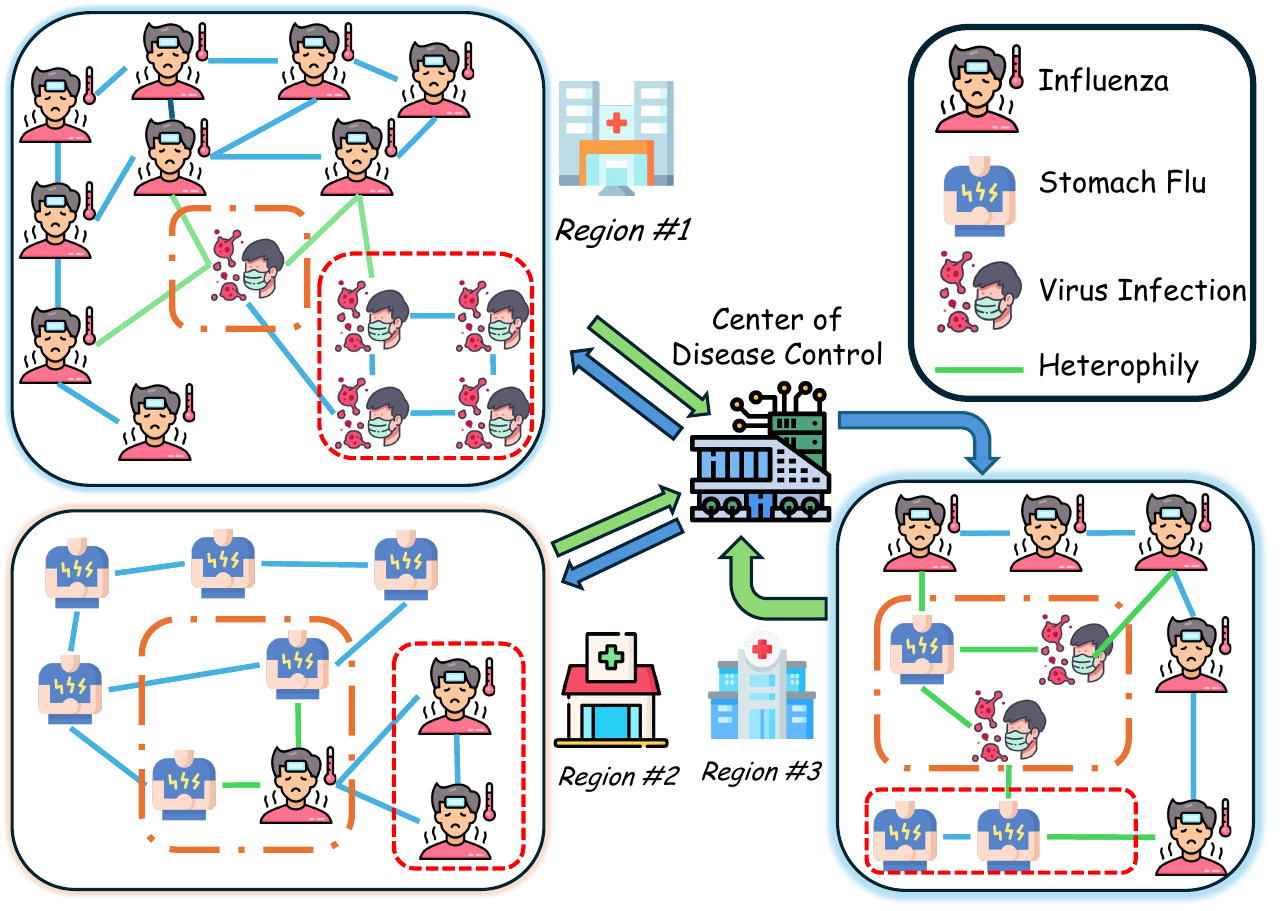}
  \captionsetup{font={small, stretch=1}}
  \caption{
    Collaborative pandemic analysis across regions with edge representing social interactions between patients.
}
\vspace{-0.1cm}
  \label{fig: realistic scenarios}
\end{figure}

    To achieve effective FGL, addressing the unique Data Heterogeneity (DH) is critical.
    The complexity of this DH stems from 
    (1) heterogeneous statistical distribution across clients (e.g. node labels and features) and 
    (2) diverse topology of subgraphs held by each client, including homophilous connectivity (i.e., connected nodes share similar labels) and heterophilous connectivity (i.e., connected nodes possess dissimilar labels and potentially misleading features~\cite{AdaFGL,FedGTA}). 
    While current FGL studies primarily aim to mitigate DH's negative impacts and improve overall predictive performance, they generally lack fine-grained explorations on node groups with disadvantageous properties and fail to thoroughly assess whether models exhibit equitably robust performance across them.

    We observe that certain nodes in FGL settings—specifically those with limited local class-wise information or pronounced heterophilous connectivity—tend to acquire suboptimal representations due to the limitations of the message-passing mechanisms of local models.
    These issues are further exacerbated by cross-client heterogeneity, where the default server-side aggregation strategies fail to adequately compensate for structural and statistical disparities across clients. Collectively, these observations highlight a fundamental fairness concern within the DH challenge: \textit{\textbf{overall predictive efficiency does not equitably reflect the biased performance towards disadvantaged groups of nodes}}, raising critical concerns about the robustness and equity of existing FGL methods. 

    To address this, we propose a dual-objective framework for \textbf{(1) Fairness for Minority Classes (FMC)}: We treat each class as the fundamental unit of fairness and promote learning efficacy for underrepresented classes within local distributions;
    \textbf{(2) Fairness for Heterophilous Nodes (FHN)}: We seek to mitigate topological bias by improving the representational quality for nodes with heterophilous connectivity patterns.
    
    This issue is especially pronounced in real-world scenarios (Fig.~\ref{fig: realistic scenarios}), where each regional institution exhibits epidemic patterns dominated by its local majority class—consistent with the homophily assumption~\cite{Homophily_Assumption}, which posits that connected nodes tend to share similar labels or features. 
    Consequently, local models exploit this structural bias to achieve strong performance in majority classes. 
    At the global level, default server-side aggregation further favors the global majority disease—such as Influenza—while underrepresented local minority classes like Virus Infections in Region~\#1 or Stomach Flu in Region~\#3. 
    These minority diseases, constrained by limited local data and heterophilous structures, are underrepresented in the aggregated model, leading to misdiagnoses of rare pathologies and reduced effectiveness in regions where minority diseases prevail.

    During our investigation, there are few FGL studies that explicitly address fairness issues.
    However, recent personalized FGL approaches~\cite{Fedspray,O-PFGL} aim to enhance local FMC by estimating global class-wise structural proxies or aligning feature distributions. 
    Another line of work~\cite{FedSig} addresses bias by generating synthetic nodes with minority labels to align client objectives.
    Despite their effectiveness, they introduce topological distortion (i.e., neglect FHN) and alter local feature distribution. 
    \textbf{\textit{In this work, we aim to develop a generalizable framework for FMC and FHN that learns from heterogeneous clients while preserving inherent data distributions and maximizing the utility of multi-client collaboration.}}
    
    To provide a clear and intuitive illustration, we conduct a detailed empirical analysis (Sec.~\ref{sec: Empirical Studies}) of existing representative FGL methods from a fairness perspective. 
    The acquired conclusion reveals the importance of evaluating fairness at a fine-grained node level, where class imbalance and topological heterogeneity must be explicitly modeled and assessed. 
    Moreover, the acquired insight emphasizes the cross-client transfer of reliable, structure-informed knowledge as a means to enable more equitable learning outcomes across diverse node groups in the FGL framework. 
    These takeaways inform the design principles of key modules in our proposed approach, enabling their coherent and synergistic integration.
    
    Specifically, motivated by the above empirical analysis, we introduce a Fairness-centered FGL framework, FairFGL, that supervises local training through the fine-grained lens on nodes' inherent properties and promotes fair subgraph-FL training through the novel collaborative mechanism.
    In client-side training, promoting both FMC and FHN is achieved by preventing local models from overfitting to dominant node groups and by integrating class-wise, topology-reliable knowledge shared from the server.
    The \textbf{History-Preserving Module} stabilizes global updates, regulating local training and transferring global knowledge to the local training model via knowledge distillation. 
    The \textbf{Majority Alignment Module} addresses topological heterogeneity by correcting biases in majority-class nodes that arise due to heterophilous structures. Aligning majority-class representations further enhances the trustworthiness of local knowledge in the collaborative learning process.
    Meanwhile, the \textbf{Gradient Modification Module} injects minority-class knowledge derived from a tailored client that possesses abundant, homophilous-connected samples of the minority classes present in the recipient client, thereby mitigating the degradation of minority-class representations induced by heterophilous connectivity.
    On the server side, training is designed to support client-side modules. Instead of transmitting full model parameters, FairFGL shares only the \textbf{Top-$k$ most influenced parameters} from local training updates, which effectively reflect local distributional characteristics while reducing both communication overhead and interference from redundant noise. 
    To address global majority dominance and reconcile conflicting local knowledge, a \textbf{Clustering-based aggregation mechanism} is employed. 
    Moreover, FairFGL selects and transmits the tailored Deviated Model reflecting the least similar data distribution to the recipient client to support the Gradient Modification Module in the subsequent training.
    
    In summary, our main contributions are presented as follows: 

    \noindent
    (1) \underline{\textbf{New Perspective}}: We broaden the scope of existing discussions on class-imbalanced scenarios by formally introducing a systematic definition of fairness learning in subgraph-FL, extended to incorporate topological properties. This formulation establishes a foundational framework for future research aimed at enhancing the robustness of FGL methods—an aspect that is particularly critical for real-world deployments. 

    \noindent
    (2) \underline{\textbf{New Paradigm}}: This paper introduces FairFGL, a novel FGL framework motivated by empirical analysis. It comprises carefully designed modules guiding fairness-aware local training, which is reinforced by a novel collaborative mechanism that facilitates reliable knowledge transfer across clients. 

    \noindent
    (3) \underline{\textbf{State-of-the-Art Performance}}: Experimental results demonstrate that FairFGL achieves state-of-the-art fairness performance across diverse cross-domain datasets. It not only improves overall Macro-F1 scores by up to 8.07\% but yields more substantial gains up to 22.62\% for three disadvantaged node groups, all while exhibiting the fastest convergence.

\section{Preliminaries}

\label{preliminary}
This section presents the key notations and formal problem definitions that form the foundation of our study. It also reviews existing FGL approaches, highlighting their limitations in addressing the dual objectives of fairness-aware optimization. Notably, the Fairness Challenge defined in this work departs from conventional fairness literature, which primarily targets bias arising from sensitive features. Instead, our formulation is tailored to the FGL context, emphasizing structural and distributional disparities across decentralized graph data.

\subsection{Problem Formulization}

Consider a graph $G = (\mathcal{V}, \mathcal{E})$  with $|\mathcal{V}|=n$ nodes and $|\mathcal{E}|=m$ edges,  the adjacency matrix (including self-loops) is $\hat{\mathbf{A}}\in\mathbb{R}^{n\times n}$, the feature matrix is $\mathbf{X} = \{x_1,\dots,x_n\}$ in which $x_v\in\mathbb{R}^{f}$ represents the feature vector of node $v$ , and $f$ represents the dimension of the node attributes. Besides, $\mathbf{Y} = \{y_1,\dots,y_n\}$ is the label matrix, where $y_v\in\mathbb{R}^{|\mathbf{Y}|}$ is a one-hot vector and $|\mathbf{Y}|$ represents the number of the classes. 
This work considers a semi-supervised node classification task, where a model trained on a labeled node set $\mathcal{V}_L$ leverages the graph topology on $\mathcal{V}_L$ and $\mathcal{V}_U$ to predict labels for the unlabeled nodes in $\mathcal{V}_U$.
Conventional FGL studies predominantly adopt \textit{Accuracy} as the evaluation metric. However, Accuracy can be misleading in settings with imbalanced class distributions, as it is often dominated by the performance on well-learned, homophilous majority classes. To address this limitation, we employ the \textit{Macro-F1} metric, which assigns equal importance to each class and better reflects a model’s ability to learn minority classes. The Macro-F1 score is defined as:
\begin{equation}
    \label{Macro-F1}
    \text{F1}_y = \frac{2 \cdot \text{Precision}_y \cdot \text{Recall}_y}{\text{Precision}_y + \text{Recall}_y}, 
    \;
    \text{Macro-F1} = \frac{1}{|\mathbf{Y}|} \sum_{y=1}^{\mathbf{Y}} \text{F1}_y,
\end{equation}
here, $\text{Precision}_y$ measures the correctness of predictions for class $y$, and $\text{Recall}_y$ quantifies the model’s ability to identify all nodes belonging to class $y$. $\text{F1}_y$ is the harmonic mean of precision and recall, and $\text{Macro-F1}$ is the unweighted average of F1 scores across all $|\mathbf{Y}|$ classes. For consistency, we refer to Macro-F1 as \underline{F1} throughout this paper since each defined node group contains multiple classes.

\subsection{Prior GNNs in Centralized Settings}
Graph convolution-based models, inspired by spectral graph theory and deep neural networks, were first introduced by ~\cite{2013firstgnn}. However, their practicality is limited by the high computational cost associated with eigenvalue decomposition. To overcome this challenge, Graph Convolutional Networks (GCNs) ~\cite{kipf2016gcn} adopt the widely accepted homophily assumption in graph learning. By using a first-order approximation of Chebyshev polynomials, GCNs simplify the topology-based convolution operator. During training, they iteratively propagate node attribute information to adjacent nodes, demonstrating strong performance in label prediction tasks.
    The forward propagation process of the $l$-th layer in GCN can be formulated as
    \begin{equation}
        \label{eq:gcn}
        \mathbf{X}^{(l)} = \sigma\left(\tilde{\mathbf{A}}\mathbf{X}^{(l-1)}\mathbf{W}^{(l)}\right),\;\tilde{\mathbf{A}} = \hat{\mathbf{D}}^{r-1}\hat{\mathbf{A}}\hat{\mathbf{D}}^{-r},
    \end{equation}   
where $\hat{\mathbf{D}}$ denotes for its degree matrix of $\hat{\mathbf{A}}$, $\mathbf{W}$ represents the trainable weights matrix, and $\sigma(\cdot)$ represents the non-linear activation function. $r\in[0,1]$ denotes the convolution kernel coefficient, which determines the extent of information flow between connected nodes during the propagation. Recent study~\cite{atp} examines the underlying correlation between the value of r and its propagation efficiency in different topologies. In GCN, by setting $r=1/2$, we can acquire $\hat{\mathbf{D}}^{-1/2}\hat{\mathbf{A}}\hat{\mathbf{D}}^{-1/2}$, known as symmetric normalized adjacency matrix, which treats both the node and its neighbors' attribute evenly during the propagation. Based on such settings, Some works~\cite{hamilton2017graphsage,2019appnp,wang2020gcnlpa,chen2020gcnii} achieve decent performance on model architectures. 
    
\subsection{Federated Graph Learning}
FGL faces an unique topological heterogeneity challenge due to the inter-dependencies between connected nodes in graph datasets. For a simplified illustration, we pair 2-layers GCN as the backbone model with FedAvg~\cite{mcmahan2017fedavg} as default aggregation strategy at the server. The training process is defined as:  
    \begin{equation}
    \label{GCN model training with parameters}
        \begin{aligned}
             \mathbf{W}_{i}^{t} = \operatorname{arg} \underset{\mathbf{W}}{\min} \frac{1}{|{\mathcal{D}_i}|} \sum_{(x,y)\in{\mathcal{D}_i}} \ell(f_{\theta}(x), y),
        \end{aligned}    
    \end{equation}
where $f_{\theta}$ is the trained model parameterized by $\mathbf{W}$,  $y\in\mathbf{Y}$, $\mathcal{D}_i$ is the local dataset of $i$-th client, and $\ell$ is the loss function fits for down-stream tasks. After the completion of local trainings, each client upload its locally trained parameters to the server for aggregation, and FedAvg follows the weighted strategy proportional to the sample size of each local client:  
    \begin{equation}
    \label{Fedavg formula}
        \begin{aligned}
            {\mathbf{W}}_{global}^{t} = \sum_{i=1}^{\mathbf{N}} \frac{n_i}{n} {\mathbf{W}_i^t},
        \end{aligned}
    \end{equation}
in which $n_i$ represent the sample size of $i$-th client, and $n$ is the total data size. $\mathbf{W}_{global}^{t}$ represents the aggregated global parameter at the server. Then, the server will broadcast the global model of round $t$ to the local client as the initiation point at the local training of next communication round.
In recent years, several subgraph-FL methods have been proposed to address the distribution heterogeneity (DH) challenge. FGGP~\cite{FGGP} utilizes well-generalized class-wise prototypes enriched with domain semantics to facilitate training across clients from diverse domains. FGSSL~\cite{FGSSL} enhances discriminability by aligning local and global representations through contrastive learning and distills structural knowledge using similarity distributions between adjacent nodes. Fed-PUB~\cite{baek2022fedpub} clusters local updates by computing similarities between local functional embeddings, thereby avoiding the aggregation of incompatible knowledge on the server. FedDEP~\cite{feddep} introduces a deep neighbor generator that leverages GNN embeddings to recover missing neighbor information while incorporating a privacy-preserving protocol based on noiseless edge-local differential privacy. FedTAD~\cite{fedtad} employs topology-aware prototypes as trustworthy semantic carriers to improve performance on node classification tasks.
%Despite advances in efficiency, current approaches leave room for more robust solutions. Most existing studies define biased nodes primarily from a class-centric perspective, overlooking the critical role of topological characteristics in graph-based prediction. From a methodological standpoint, synthetic methods risk distorting local topology and introducing noise through inaccurate edge generation. Personalized approaches often lack generalizability across unseen scenarios. Incentive-based mechanisms may induce inequitable training dynamics by disproportionately rewarding a subset of clients.

\begin{figure*}[t]   %注意，这里设置是关键
	\centering
    \includegraphics[width=\linewidth,scale=1.00]{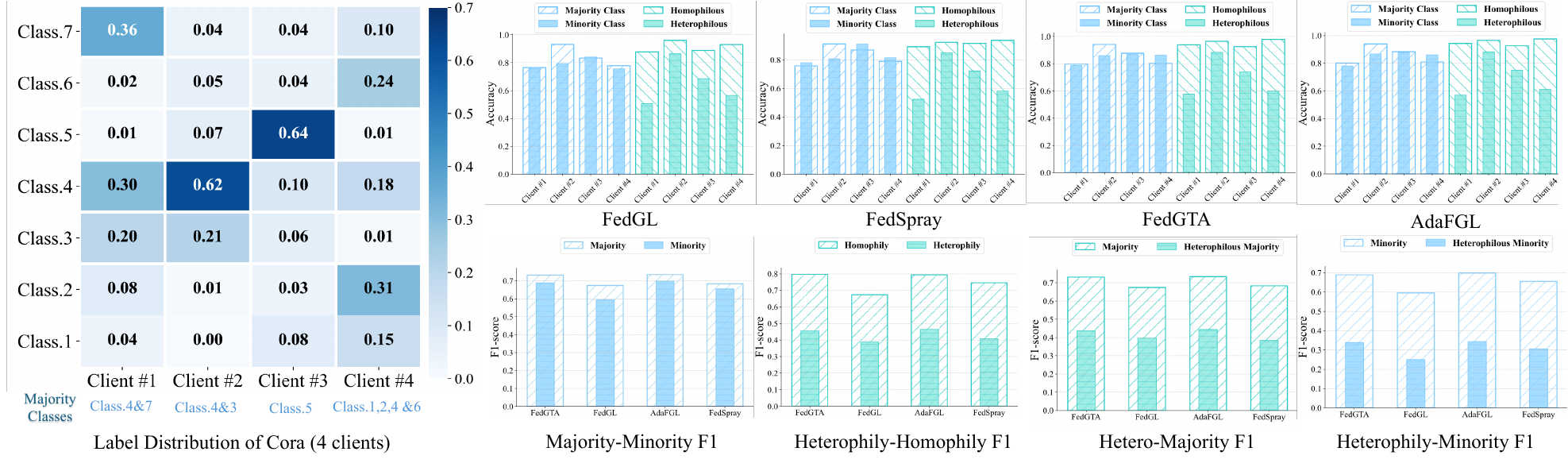}
  \captionsetup{font={small, stretch=1}}
	%[]里面的参数自己可根据需要调整
	\caption{
       Nodes volume of each class within each client is presented proportionally with indications of majority classes at the bottom. For fair comparison, we rank nodes by homophily score, selecting the top 50\% as homophilous and the rest as heterophilous. The right-hand side showcases performances of four representative FGL methods: 1) Top-Side: Individual method performance on Cora under dual-perspective partitioning strategies; 2) Bottom-side: Comprehensive Evaluation concerning Heterophilous (Hete) nodes particularly.
 }
\label{fig: empirical studies}
\vspace{-0.2cm}
\end{figure*}

\subsection{Conventinal Fairness in FL}
%Federated Learning (FL) hinges on its distributive training architecture to  mitigates the impact of heterogeneous data distribution by integrating of complementary knowledge from each independent client and recent studies have devoted in addressing issues of missing or underrepresented classes. 
%One common approach to achieving this is by modifying the loss function during the client-side training. Methods like~\cite{BalanceFL} introduce a re-weighting mechanism on the client-side to assign weights to each local class inversely proportional to its frequency in the local data, which simulates training on a uniform class distribution. 
%Others~\cite{FedeAMC,fairness_factorization} adopt similar strategies, either focusing on underrepresented classes or minimizing optimization disparities between classes.
%On the server-side, client sampling strategies~\cite{flfaultD,CUCB,Fed-Focal_Loss} have been adjusted to select clients with more balanced data distributions, thereby ensuring more class-balanced global model aggregation.
%Furthermore, some studies~\cite{FRAug,Astraea} paid efforts in data-centric perspective by employing data augmentation techniques to generate synthetic nodes or embeddings and mitigate biased data distribution. 
%Despite showcasing their efficiency, FL methods are not originally applied to the graph data type and thus lack consideration for the role of topology, which is intrinsically impactful in this work's class-imbalanced scenario. 
Federated Learning (FL) leverages its distributed training architecture to mitigate the effects of data heterogeneity by integrating complementary knowledge from decentralized clients. Recent efforts have addressed the challenge of missing or underrepresented classes~\cite{fairness_FL_Survey}.
A common approach involves modifying the client-side loss function. For instance,~\cite{BalanceFL} introduces class re-weighting based on inverse class frequency to approximate uniform class distributions. Similar strategies~\cite{FedeAMC,fairness_factorization,FAPL} focus on either enhancing underrepresented class performance or reducing inter-class optimization disparities.
On the server side, client sampling techniques~\cite{flfaultD,CUCB,Fed-Focal_Loss} prioritize clients with more balanced data to improve global aggregation fairness. Additionally, data-centric methods~\cite{FRAug,Astraea} employ data augmentation to generate synthetic nodes or embeddings, addressing class imbalance at the data level.
While effective, these FL methods are not inherently designed for graph data and thus overlook topological factors—an essential consideration in class-imbalanced graph scenarios.

\subsection{Limited Fairness in FGL}

Several recent studies have begun to address the DH challenge caused by class-imbalanced distribution at the local level, aligning to some extent with dual-fairness-centered optimization objectives. Some of them draw inspirations from the success of centralized graph learning and extend them to the FGL settings. 
FedSig~\cite{FedSig} extends the idea of node synthesis from GraphSMOTE~\cite{GraphSmote} by generating synthetic minority-class nodes through cross-client collaboration. This approach preserves each client's original data distribution while maintaining embedding proximity between local and global minority nodes. 
FedLog~\cite{FedLog} constructs a globally synthetic dataset on the server for centralized training, alleviating local overfitting and enhancing generalization under non-IID settings.
FedSpray~\cite{Fedspray} explore reliable class-wise discriminators from local majority classes enjoying the high homophily and collaboratively transfer them for local personalized training. 
FGPL~\cite{FGPL} constructs a global prototype by aggregating class-wise prototypes trained on local clients. Additionally, several methods~\cite{Pan2023TowardsFG, NoPrejudice, PFedNC} introduce incentive mechanisms that encourage fairness in model aggregation. These frameworks reward client updates that improve generalization while penalizing those that exacerbate representational bias.

\section{Empirical Studies}
\label{sec: Empirical Studies}

To further explore whether representative FGL methods fall short of achieving fairness objectives despite their reported predictive performance, we conduct the following empirical study. The \textbf{Conclusion} of this study strengthens our research motivation, and the acquired key \textbf{Insight} guides the design of FairFGL. We investigate two questions aligned with the previously defined dual-objective standards: \\
\indent \textbf{Q1} (\textbf{Fairness-centered}): Are node groups with varying properties being treated equitably by existing FGL methods?\\
\indent \textbf{Q2} (\textbf{FMC and FHN-centered}): Which properties lead to severe biased learning issues imposed on node groups?

To simulate realistic and challenging FGL scenarios, we partition the Cora dataset into four local clients using the Metis algorithm, ensuring heterogeneous label distributions across clients (see Fig.~\ref{fig: empirical studies}). We define majority classes as those whose node counts exceed the average number of samples per class. To quantify topological properties, we compute the homophily score (Eq.~\ref{eq:prototype-gen}) as the proportion of a node’s 1-hop neighbors that share the same class label. Since Cora is a highly homophilous dataset, we balance the number of homophilous and heterophilous nodes within each client to enable reliable comparisons. For evaluation, we report both Accuracy and F1-score; the latter is particularly sensitive to misclassifications of minority classes and thus provides a more robust measure of model performance across diverse node groups.

\textbf{1-Observation}: To address Q1, we evaluate four representative FGL methods by examining node performance from both class-wise and topological perspectives. In clients where majority classes are highly dominant—for example, Client 2, where they constitute over 80\% of the nodes—existing FGL methods consistently exhibit a noticeable accuracy disparity between majority and minority groups, which becomes more pronounced in F1. Among the evaluated methods, FedSpray~\cite{Fedspray} shows slight advantages in improving predictive accuracy for minority nodes, owing to its design that facilitates the transmission of reliable class-wise structural proxies between clients. When node performance is analyzed through a topological lens, the gap between homophilous and heterophilous nodes becomes significant and is further amplified under the F1 metric.

\textbf{1-Analysis}: This performance disparity originates from the intrinsic bias of message-passing neural networks, which favor homophilous structures. Majority-class nodes often form dense local communities that reinforce similar labels and acquire optimized decision boundaries. In contrast, sparsely connected or heterophilous nodes—frequently associated with minority classes—experience limited signal propagation, leading to greater representation errors.
FedSpray exhibits suboptimal performance compared to recent methods, such as FedGTA~\cite{FedGTA} and AdaFGL~\cite{AdaFGL}, that mitigate topological heterogeneity. Specifically, its reliance on global structural proxies to regulate local presentation learning lacks exploration on each nodes' topological connectivity.
\begin{figure*}[t]
  \includegraphics[width=\textwidth]{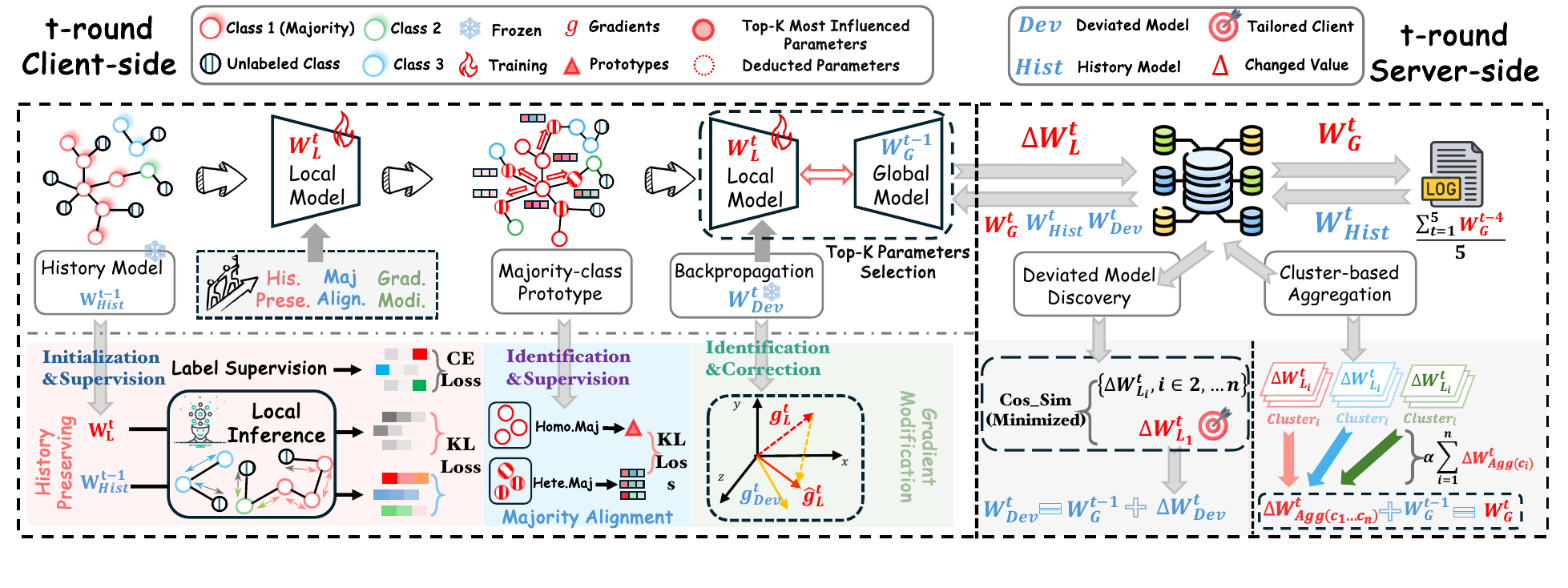}
  \vspace{-1cm}
  \captionsetup{font={small, stretch=1}}
  \caption{
  Describe the pipeline of FairFGL's training procedure. $t$ indicates the $t$-th training round. \textbf{\textit{History-Preserving Module}} leverage the received History Model to regulate model training; \textbf{\textit{Majority Alignment Module}} rectifies the Hete-Maj nodes through acquired class-wise prototype via KL loss function; \textbf{\textit{Gradient Modification Module}} rectifies the gradients directionalities of local model.}
  
  \label{fig: framework}
  \vspace{-0.4cm}
\end{figure*}
FedGTA performs personalized aggregation by prioritizing clients with similar data distributions and assigning greater weight to those with higher homophily scores. AdaFGL adapts local graph structures using global guidance and modifies propagation rules based on local homophily. While both methods outperform baselines lacking explicit topological modeling, they suffer from notable limitations. FedGTA tends to form a ``majority-class closed loop'', reinforcing dominant patterns and sidelining minority nodes. AdaFGL focuses disproportionately on densely connected regions during global topology reconstruction, limiting its ability to represent structurally sparse minority nodes.
Overall, both approaches prioritize mitigating propagation inconsistencies caused by topological heterogeneity, yet overlook node-level treatment from a micro perspective. These limitations are clearly reflected in the performance on the Heterophily-Homophily F1. 
Observation and subsequent analysis above reveals a critical perspective that learning biases largely originate from properties that shape models' efficacy in learning node representation.

\textbf{2-Observation}: To answer Q2, we conduct a fine-grained analysis to identify which node properties most hinder representation learning in graph models. As illustrated in Fig.~\ref{fig: empirical studies}, class-wise partitioning leads to relatively modest F1-score disparities. In contrast, topology-based partitioning reveals significantly larger performance gaps, underscoring topology as a key factor in ensuring fairness and robustness in FGL. This is because the topological property profoundly affects the representational quality of individual nodes. However, existing FGL studies have yet to thoroughly investigate how diverse node-level topological properties contribute to these disparities.

\textbf{2-Analysis}: This discrepancy stems from the distribution of class labels within topological groups. In homophilous groups—particularly in datasets like Cora—nodes typically belong to the majority class (Homo-Maj), aligning with the homophily assumption and yielding high F1 scores. However, this strong performance can obscure the poor outcomes experienced by majority-class nodes with heterophilous connections (Hetero-Maj). As reported in the Hetero-Majority F1, these nodes suffer from impaired representations and fail to benefit from well-learned decision boundaries, despite belonging to the majority class.
Similarly, the performance degradation of minority-class nodes with heterophilous connections (Hete-Min) is masked by the overall performance of minority nodes, some of whom possess homophilous connectivity raise the overall F1 score. These findings underscore the importance of topological context in fairness-aware learning: Unfavorable topologies amplify misclassification risks, especially for underrepresented classes.

One promising direction is to transfer high-confidence representations from well-performing Homo-Maj nodes to support Hetero-Maj nodes, whose local representations are less reliable. For minority nodes, cross-client collaboration becomes essential. A node that is a minority in one client may be a majority in another—for example, Class 5 is dominant in Client 3 but underrepresented in Clients 1 and 2.

Exploring the two questions above, we reach the following conclusion and insight. Both motivate our research on FGL's Fairness Challenge and guide the design of FairFGL:\\ 
\textbf{Conclusion}: \textit{Global accuracy alone is insufficient to capture the nuanced performance of FGL methods across diverse node groups. Metrics such as F1 and specific evaluations on varying node groups are essential for exposing model biases; fairness in FGL cannot be achieved without evaluating class imbalance and topological heterogeneity at the fine-grained node level.}\\
\textbf{Insight}: \textit{Performance degradation among nodes with unfavorable topological conditions—particularly heterophilous minority nodes—highlights the need for cross-client transfer of reliable, structure-informed representations to mitigate representational inequality.}

\section{FairFGL Architecture}
\label{FairFGL architecture}
In this section, we present method \textbf{FairFGL} that promotes Fairness Learning in FGL under the severe class-imbalance settings across clients through the collaborative mechanism.

In Section IV-A, we present the overview of FairFGL's architecture as depicted in Fig.~\ref{fig: framework} followed by the intuitive descriptions for each module that contributes to its effectiveness. Following subsections then offer detailed but intuitive interpretations on modules' design and separate them into Client-side and Server-side training to offer a clear FGL procedure. 

\subsection{Architecture Overview}
\label{Architecture overview}
FairFGL adopts aforementioned \textbf{Insight} as the guiding principle, and aims to achieve dual-objective optimization at both client and server-side procedures:

\subsubsection{Client-side Training}
We acknowledge the presence of severe class imbalance, which causes locally trained models to exhibit biased performance favoring majority classes in the absence of targeted intervention. At the local side, we first introduce the \textbf{History-Preserving Module} that averages the five most recent global updates into a uniform model. The resulted \textbf{History Model} remains frozen through local training and aids Fairness-centered objectives in two perspectives: (1) being integrated into active local training model using a trainable weight matrix to prevent the latter from favoring local majority classes; (2) transferring the global knowledge through labeling unlabeled nodes via knowledge distillation. 

Based on the \textbf{Insight}, we introduce the \textbf{Majority Alignment} mechanism specifically for improving representation learning for heterophilous majority nodes during the local training phase, enhancing local models' reliability in transferring class-wise knowledge collaboratively. While \textbf{History-Preserving Module} mainly functions as supplementing the reliable global class-wise knowledge, it is critical for acknowledging the correlation between class and topology in offering fine-grained graph knowledge. Therefore, the \textbf{Gradient Modification Module} refines model updates by incorporating knowledge from clients with dissimilar data distributions. This mechanism enhances training on underrepresented minority classes by leveraging well-labeled data and reliable topological structures from a \textbf{tailored Deviated Client}, where the minority class of the receiving client appears as a majority in its subgraph.

\subsubsection{Server-side Training}
Instead of transmitting full model parameters, FairFGL only shares changes of \textbf{Top-$k$ most influenced parameters}. This approach captures essential local data characteristics while reducing communication overhead and noise interference. To mitigate the dominance of global majority classes—those overrepresented across clients—FairFGL modifies the standard FedAvg strategy by clustering local updates and aggregating them with equal weight per cluster. This promotes a more fairness-aware global model update. Furthermore, to support the \textbf{Gradient Modification Module} in the next training round, the server identifies the least similar local update for each client. This tailored local update is then integrated with the previous round’s global model to acquire the \textbf{tailored Deviated Model} for transmission.

\subsection{Local Fairness-enhanced Learning}
\label{sec: Local-side}
\subsubsection{History-Preserving Module}
\label{subsec:History Model}
The received most recent global model becomes the initial model before local training commerce, and addition to it, each client will acquire an History Model, $\mathbf{W}_{Hist}^{t}$, from averaging the five most recent global models, which will remain frozen for the remainder of the training. Compared to the default procedure of most FGL studies that usually only utilize the most recent global update, such a design provides more enriched and consistent global knowledge. Additionally, it builds on the presumption that global models from remote rounds might be too divergent from the current training to contribute any valuable improvements. 

%The newly received global model from the previous round replaces the previously trained local model as $\mathbf{W}_{Local}^{t}$. 

The History Model involves in client-side training on two levels: (1) it is fused with locally trained parameters using learnable weights $\alpha$ as an adaptive regularization term; (2) it functions as a teacher model to transfer global knowledge to the local model on the unlabeled set, $\mathcal{V}_U$, via knowledge distillation, enriching the training data with reliable global minority knowledge. The process is defined as follows: 
    \begin{equation}
    \label{eq:Local Ce_Loss}
        \begin{aligned}  
        \hat{\mathbf{W}}_{local}^{t} & = {\mathbf{W}}_{Hist}^{t}\times \mathcal{\alpha}+{\mathbf{W}}_{local}^{t}\times(1-\mathcal{\alpha}),
        \end{aligned}
    \end{equation}
    \begin{equation}
    \label{eq:kd_for_Unlabeled}
    \mathcal{L}_{\text{Distill}} 
    = \frac{1}{|\mathcal{V}_U|} \sum_{i \in \mathcal{V}_U} 
    {\rm KL}\left( 
        \mathrm{softmax}\left(\mathbf{z}_{\text{hist}}^{(i)} \,\middle\|\,\mathbf{z}_{\text{local}}^{(i)}\right)
    \right), 
    \end{equation}
where $\mathbf{z}_{\text{hist}}^{(i)}$ and $\mathbf{z}_{\text{local}}^{(i)}$ denote as logits outputs of unlabeled nodes predicted by the History Model and local training model, respectively. $\mathrm{KL}$ denotes for the KL divergence loss function. Unlike synthetic approaches that generate nodes to address class-imbalanced distribution, this module leverages client-specific knowledge while preserving the original distribution.

\subsubsection{Majority Alignment Module}
\label{sec: Majority Alignment Module}
Acquired \textbf{Insight} underscores the significant decline in performance observed in nodes exhibiting disadvantageous heterophilous connections, but also delineates the potential of applying topology-driven rectification for heterophilous majority nodes during the local training phase due to their ample local sample size. This module improves predictive reliability for majority-class nodes, thereby reinforcing the efficacy of the knowledge transfer mechanism across clients. This enhancement benefits both the subsequent Gradient Modification Module and the aggregation process at the server. Such a process can be defined as follows:
    \begin{equation}
    \label{eq:prototype-gen}
    \begin{aligned}
        \mathcal{H}_{v_i} &= \frac{\sum_{v_j \in \mathcal{N}(v_i)} \mathbb{I}(y_j=y_i)}{|\mathcal{N}(v_i)|},
        \;
        \mu_y = \frac{\sum_{v_i \in \mathcal{V}_y^{\text{Homo}}}h_{i}}{|\mathcal{V}^{\text{Homo}}_{y}|},
\end{aligned}
\end{equation}
\begin{equation}
\label{eq:stru_loss}
    \begin{aligned}
    \mathcal{S}_{loss} 
    &= \frac{1}{ |\mathcal{V}_{\text{y}}^{\text{Hete}}|} \sum_{v_i \in \mathcal{V}^{\text{Hete}}_{\text{y}}}
    {\rm KL}( \mathrm{softmax}\left(h_i \,\middle\|\, \mu_{y}\right)).
    \end{aligned}
\end{equation}

\begin{algorithm}[t]
\caption{Client-side Training (Sec.~\ref{subsec:History Model} to Sec.~\ref{sec: top-k parameter selection})}
\label{algorithm 1: Local model training}
\begin{algorithmic}[1]  % the [1] allows line numbers for the following procedure
    \STATE \textbf{Definition.} $g_{local}$ is the gradient of local training model, $g_{dev}$ is the Deviated Model's gradient, $M_{dev}$ locates subset of parameters that will be modified.\\
    \FOR {each communication round $t = 1, ..., T$}
    \FOR {\textbf{parallel} each client $n = 1, ..., N$}
    \STATE \textbf{Received:} $\mathbf{W}_{global}^{t}$, Acquire $\mathbf{W}_{Hist}^{t}$,\\
    \FOR {each local model training epoch $e = 1, ..., E$}
    \STATE Conduct local training by Eq.~\ref{eq:Local Ce_Loss} - ~\ref{eq:stru_loss};\\
    \STATE \textbf{parallel Execute} Gradient Modification Module (Sec.~\ref{sec:gradient_mod}):\\
    \STATE  ${d} = {g}_{local} \cdot {g}_{dev}$,\;\;\;\;\;\;\;\;\;\;\;\; \textit{Proceed if}\;$d$ \textit{is negative}\\
    \STATE ${g}_{projected} = \frac{{d}}{{\Vert {g}_{dev}\Vert}^{2}}\times{g}_{dev}$,\\
    \STATE ${g}_{corrected} = {g}_{local} - {g}_{projected}$,\\
    \STATE \textbf{Follow} the Binary Law illustrated at Eq.~\ref{eq: binary law} to update the subsets of ${\mathbf{W}}_{local}^{t}$ with $M_{dev}$. \\
    \ENDFOR
    \STATE Execute Top-$k$ parameters Selection by Eq.~\ref{eq:top-k-parameter-selection} - ~\ref{eq: top-k 2};\\
    \STATE Fine-tuning set of the ${\Delta{\mathbf{W}}_{local}^{t}}$ reflected the same parameters' location indicated in ${\Delta{\tilde{\mathbf{W}}}_{local}^{t}}$;\\
    \STATE \textbf{The final trained local model:}\\ $\hat{\mathbf{W}}_{local}^{t}\gets{\mathbf{W}}_{Hist}^{t}\times \mathcal{\alpha}+{\mathbf{W}}_{local}^{t}\times(1-\mathcal{\alpha})$.\\
    \ENDFOR
    \STATE \textbf{Each Client:} Upload the Fine-tuned ${\Delta{\tilde{\mathbf{W}}}_{local}^{t}}$ to Server.\\
    \ENDFOR
\end{algorithmic}
\end{algorithm}
\noindent
$\mathcal{H}_{v_i}$ indicates the homophily score calculation, which is applied for each majority node. Based on results, the majority group is then partitioned into Homo-Maj nodes and Hete-Maj groups. We then acquire the prototype of majority class $y$ from $\mathcal{V}^{\text{Homo}}_{y}$, and distill its knowledge to $\mathcal{V}_{\text{y}}^{\text{Hete}}$ in Eq.~\ref{eq:stru_loss}.

\subsubsection{Gradient Modification}
\label{sec:gradient_mod}
Since the server-side aggregation process risks shifting the global optimum toward the global majority class, acquiring sufficient minority-class knowledge tailored to each client is crucial. While existing methods employ synthetic techniques or loss function constraints to augment local minority-class knowledge, FairFGL leverages the collaborative nature of FGL alone.

On the server side, we introduce similarity-guided pairing, where each client is matched with another possessing complementary expertise in minority classes to mitigate overfitting to local label skew. This process produces a tailored \textbf{Deviated Model} for each client, delivered by the server. 
Minority expertise of the Deviated Model is enriched by ample data samples that are structurally homophilous, and is carried within its trained local parameters as proxy reflecting its local data distribution. The details of selecting Deviated Model on the server-side is discussed by Eq.~\ref{eq: find deviated model}.

We designate this tailored Deviated Model for the $i$-th client in the $t$-th training round as $\mathbf{W}_{i_{dev}}^{t}$. The Parameter-Activation Mask $\mathcal{M}_{dev}$ is applied to locate the parameter subset that best encapsulates the unique data characteristics of the selected client, and activates only such a subset. During local training, gradient descent is adapted to integrate minority-class knowledge by refining gradient updates in backpropagation. We provide detailed formulas in Algorithm~\ref{algorithm 1: Local model training} and as follows:
\begin{equation}
\label{eq: binary law}
  \hat{g}_{\mathrm{local}} =
  \begin{cases}
    g_{\mathrm{local}}, & \text{if } d \geq 0 \\
    g_{\mathrm{correct}} + \frac{\mathrm{margin} \cdot g_{\mathrm{dev}}}{\lVert g_{\mathrm{dev}} \rVert}, & \text{if } d \le 0
  \end{cases}.
\end{equation}
The function in Line 9 determines whether the local training model and $\mathbf{W}_{i_{dev}}^{t}$ exhibit similar or divergent gradient directions. If the outcome is greater than 0, indicating that the gradients are aligned, we avoid disrupting local optimization. Otherwise, we proceed with gradient modification as specified by the functions in Lines 10 and 11 of Algorithm~\ref{algorithm 1: Local model training} and Eq.~\ref{eq: binary law}.
Overall, this procedure minimizes conflict between the local model and $\mathbf{W}_{i_{dev}}^{t}$, as the latter possesses more reliable minority-class knowledge. Additionally, we incorporate the $\operatorname{margin}$ as a regulatory term to preserve local knowledge.

   % \begin{numcases}
        %{\hat{g}_{local}=}
        %{g}_{local}& $\text{if}\; {d}\geq 0$ \nonumber \\
        %{g}_{correct}+ \frac{\operatorname{margin}\times{g}_{dev}}{\Vert {g}_{dev}\Vert} & $\text{if}\; {d}\le 0$.
   % \end{numcases}
\subsection{Federated Learning}
\label{sec: federated learning}
\subsubsection{Top-k Changed Parameters selection}
\label{sec: top-k parameter selection}
Instead of transmitting gradients or full model parameters to the central server, we identify and upload only the subset of parameters most impacted during the current round of local training.
The intuition is that these parameter changes capture the most informative local knowledge reflective of the client's data characteristics. Prior works~\cite{Graph_pruning, Predict_Parameters} assess parameter importance based on their absolute values. However, under severe class-imbalanced distribution, this approach tends to over-represent majority-class knowledge. In contrast, parameter changes reveal which components are actively updated during training. This process is illustrated as:
\begin{equation}
\label{eq:top-k-parameter-selection}
  \Delta \mathbf{W}_{\mathrm{local}}^{t} = \left| \mathbf{W}_{\mathrm{local}}^{t} - \mathbf{W}_{\mathrm{global}}^{t} \right|,
\end{equation}
\begin{equation}
\label{eq: top-k 2}
  \Delta \tilde{\mathbf{W}}_{\mathrm{local}}^{t} = 
  \begin{cases}
    \Delta \mathbf{W}_{\mathrm{local}}^{t} & \text{if } \Delta \mathbf{W}_{\mathrm{local}}^{t} \geq \rho \\
    0 & \text{otherwise}
  \end{cases},
\end{equation}
where $\rho$ is a quantile that decides the ratio of Top-$k$ parameters that we select as local task's knowledge, and the $\mathbf{W}_{global}^{t}$ is identical to all clients. We also provide the overview of the local training process in the Algorithm~\ref{algorithm 1: Local model training} for your reference.

\subsubsection{Fairness-enhanced aggregation mechanism}
\label{subsec: Federated Training}
Due to varying data distributions across clients, the indiscriminate adoption of FedAvg as an aggregation strategy inevitably leads to conflicts arising from incompatible knowledge, as demonstrated in prior studies~\cite{baek2022fedpub, FedGTA}. To address this issue, we develop an effective clustering process using the intuitive $k$-means algorithm, where the number of clusters is dynamically determined by the Silhouette Coefficient.

\begin{algorithm}[t]
\caption{Federated Training (Sec.~\ref{subsec: Federated Training})}
\label{algorithm 2: Federated training}
\begin{algorithmic}[1]  % the [1] allows line numbers for the following procedure
    \FOR {each communication round $t = 1, ..., T$}
    \STATE \textbf{Server Receives:} ${\Delta{\hat{\mathbf{W}}}_{local}^{l}}$;\\
    \STATE Conduct $k$-mean algorithm for clustering process based on Silhouette Coefficient;\\
    \STATE Conduct the Weighted Average Aggregation within each cluster by Eq.~\ref{eq:FedAvg-aggregation-within-cluster};\\
    \STATE Acquire the global aggregated updates by Eq.~\ref{eq:aggregation-between-clusters};\\ 
    \STATE Update the global aggregated model by Eq.~\ref{eq:update-global-model-at-server};\\
    \STATE \textbf{For each client:} Find $\mathbf{W}_{i_{dev}}^{t+1}$ by Eq.~\ref{eq: find deviated model};\\
    \STATE \textbf{Broadcast:} $\mathbf{W}_{i_{dev}}^{l+1}$, $\mathbf{W}_{global}^{l+1}$ to Clients.\\
    \ENDFOR
\end{algorithmic}
\end{algorithm}

Within each cluster—comprising clients with similar data distributions—we apply the standard FedAvg~\cite{mcmahan2017fedavg} to aggregate local updates. The aggregated cluster-level updates are then combined using a uniform learning rate to compute the final parameter updates for the current round. These updates are applied to the previous global parameters to obtain the updated global parameters for this round.
\begin{table*}[htbp]
\centering
\setlength{\abovecaptionskip}{0.2cm}
\caption{The statistical information of selected datasets.
}
\label{tab: subgraph_fl_datasets}
\resizebox{0.8\linewidth}{!}{
\setlength{\tabcolsep}{3mm}{
\begin{tabular}{ccccccc}

\toprule
Subgraph-FL       & Nodes     & Edges      & Features & Classes & Train/Val/Test & Description \\ \midrule[0.3pt] %Topological Nature~\cite{heterophilic_handbook}           \\ \midrule[0.3pt]
Cora              & 2,708     & 5,429      & 1,433    & 7       & 20\%/40\%/40\% & Citation Network \\ %& Homophilic      \\
CiteSeer          & 3,327     & 4,732      & 3,703    & 6       & 20\%/40\%/40\% & Citation Network \\ %& Homophilic      \\
PubMed            & 19,717    & 44,338     & 500      & 3       & 20\%/40\%/40\% & Citation Network \\ \midrule[0.3pt] %& Homophilic      \\ \midrule[0.3pt]
Chameleon         & 2,277     & 36,101     & 2,325    & 5       & 48\%/32\%/20\% & Wiki-page Network \\ %& Heterophilic    \\
Squirrel          & 5,201     & 216,933    & 2,089    & 5       & 48\%/32\%/20\% & Wiki-page Network \\ \midrule[0.3pt] %& Heterophilic    \\ \midrule[0.3pt]
Minesweeper       & 10,000    & 39,402     & 7        & 2       & 50\%/25\%/25\%   & Game Synthetic Network \\ \midrule[0.3pt] %& Homophilic \\ \midrule[0.3pt]
Tolokers          & 11,758    & 519,000    & 10       & 2       & 50\%/25\%/25\%   & Crowd-sourcing Network\\ \bottomrule %& Homophilic \\ \bottomrule[0.3pt]
\end{tabular}
}}
\vspace{-0.2cm}
\end{table*}
This design also addresses potential performance degradation caused by the dominance of global majority classes. Since clients where the global majority also appears as the local majority tend to share similar data distributions, we mitigate this bias by applying a uniform learning rate to each cluster-level update during aggregation. This process is formulated as follows: 
\begin{equation}
\label{eq:FedAvg-aggregation-within-cluster}
  \sum_{i=1}^{N_k} \frac{n_i}{n_k} \Delta \tilde{\mathbf{W}}_{i}^{t} 
  \;\Longrightarrow\; 
  \Delta \widetilde{\mathbf{W}}_{C_k}^{t+1},
\end{equation}
\begin{equation}
\label{eq:aggregation-between-clusters}
  \sum_{k=1}^{K} \Delta \widetilde{\mathbf{W}}_{C_k}^{t+1} \cdot \eta
  \;\Longrightarrow\; 
  \Delta \mathbf{W}_{\mathrm{Agg}}^{t+1},
\end{equation}
\begin{equation}
\label{eq:update-global-model-at-server}
  \Delta \mathbf{W}_{\mathrm{Agg}}^{t+1} + \mathbf{W}_{\mathrm{global}}^{t}
  \;=\; 
  \mathbf{W}_{\mathrm{global}}^{t+1},
\end{equation}
\noindent
in which $n_i$ represents the number of data samples in $i$-th client assigned to cluster $k$, and $n_k$ represent the total number of data samples in the $k$-th cluster.

Simultaneously, we aim to identify a tailored Deviated model for each client to support the Gradient Modification Module in the next round. To this end, we adopt a cosine similarity-based minimization objective to assign each client the local update that maximally reflects minority-class knowledge. Finally, the selected deviated update is combined with the previous global model to construct the Deviated model:
\begin{gather}
\label{eq: find deviated model}
    (i^{\ast}, j^{\ast}) = \operatorname{arg} \underset{i,j\in N}{\min}
    \frac{
        \Delta{\tilde{\mathbf{W}}_{i}^{t}} \cdot \Delta {\tilde{\mathbf{W}}_{j}^{t}}
    }{
        \Vert \Delta {\tilde{\mathbf{W}}_{i}^{t}} \Vert \cdot \Vert \Delta {\tilde{\mathbf{W}}_{j}^{t}} \Vert
    }, \\
    \mathbf{W}_{global}^{t} + \Delta \tilde{\mathbf{W}}_{j}^{t} \longrightarrow \mathbf{W}_{i_{dev}}^{t+1},
\end{gather}
where, $(i^\ast, j^\ast)$ represent the resulted pair for each client. By the end of the server-side operation, server will broadcast $\mathbf{W}_{global}^{l}$,$\mathbf{W}_{i_{dev}}^{l+1}$, and the mask indicating the location of updated parameters from the Deviated client to each local client for faciliating the next round of training, and the whole federated training will continue until the convergence has been reached. See details of the federated training process in Algorithm~\ref{algorithm 2: Federated training}.

\section{Experiment}
\label{sec: experiment}
In this section, we offer a comprehensive evaluation of FairFGL, focusing on proposed dual-objective Fairness-centered optimization framework. 
Building upon the recently developed FGL benchmark system~\cite{openfgl}, this experiment is designed to simulate severe class-imbalanced distribution both within and across local clients, as visualized in the heat map of Fig.~\ref{fig: empirical studies}. 
To achieve this, we adopt the Metis algorithm as the partitioning method. Compared to the Louvain algorithm used in prior studies~\cite{Fedspray}, Metis minimizes cross-client correlations without explicitly maximizing modularity, effectively simulating skewed label distributions across subgraphs assigned to different clients.

The evaluation metrics include the commonly used \textit{Accuracy} in FGL research and \textit{F1}, which is more sensitive to misclassifications on minority-class nodes. 
Specifically, we assess model performance across different node groups, including: Minority-class nodes (Min-F1), nodes with heterophilous connectivity (Hete-F1), and nodes that belong to intersection of both above groups (Hete-Min-F1). This categorization aligns with our optimization objectives, addressing fairness challenges from both class-wise and topology-wise perspectives.

To ensure a robust evaluation across diverse real-world applications, we include seven datasets spanning four domains in Subgraph-FL evaluations: Citation networks: Cora, Citeseer, Pubmed; Wiki-page networks: Squirrel, Chameleon; Synthetic game networks: Minesweeper; Crowdsourcing networks: Tolokers. A detailed description of each dataset, along with corresponding training configurations is presented in Table~\ref{tab: subgraph_fl_datasets}.
Specifically, we select graph datasets from the OpenFGL collection with 2 to 7 classes to ensure that each client retains a balanced representation of classes under subgraph-FL simulation, minimizing the risk of class absence in the test set.

For baseline comparisons, we evaluate FairFGL against recently representative FGL methods, including: FedSage+~\cite{zhang2021fedsage}, FedGL~\cite{chen2021fedgl}, FedPub~\cite{baek2022fedpub}, FedGTA~\cite{FedGTA}, AdaFGL~\cite{AdaFGL} and FedTAD~\cite{fedtad}. Additionally, we incorporate FL methods,including FedAvg~\cite{mcmahan2017fedavg}, Scaffold~\cite{karimireddy2020scaffold}, FedProx~\cite{li2020fedporx}, and Moon~\cite{li2021moon}, to assess how well our approach addresses fairness challenges in graphs. To ensure fair comparisons, we either adopt the hyperparameter settings recommended by the original authors of each baseline or perform automatic hyperparameter tuning using Optuna~\cite{akiba2019optuna} when recommendations are unavailable.

We set the number of local epochs per round to three for all baselines, except for methods that require additional local fine-tuning. Each experiment is run for 150 rounds of federated training, based on our observation that most methods require over 100 rounds to converge. All results are correspondent to the best performance reported on the validation set.
\begin{table*}[t]
\caption{
\centering 
The best result is in \textbf{bold}, and the second best results are \underline{underlined}. The Standard Deviations are provided in brackets.}
\label{tab: Subgraph-FL Table 1}
\resizebox{\linewidth}{25mm}{
\setlength{\tabcolsep}{2.5mm}{
\begin{tabular}{c|cc|cc|cc|cc|cc|cc|}
\midrule[0.3pt]
Method   &  \multicolumn{2}{c|}{Cora}  &  \multicolumn{2}{c|}{CiteSeer}  &  \multicolumn{2}{c|}{PubMed}   & \multicolumn{2}{c|}{Chameleon} &  \multicolumn{2}{c|}{Squirrel} & \multicolumn{2}{c|}{Minesweeper} \\ \midrule[0.2pt]
Metrics  & Overall F1  & Accuracy  & Overall F1  & Accuracy & Overall F1  & Accuracy & Overall F1  & Accuracy & Overall F1  & Accuracy & Overall F1  & Accuracy    \\ \midrule[0.3pt]
FedAvg   & 53.70(0.30)    & 81.13(0.33)      & 53.37(0.74)     & 68.74(0.35)    & 62.82(0.15)    & 84.80(0.01)    & 50.93(0.88)   & 62.14(0.70)  & 43.07(0.80) & 44.91(1.35) & 49.52(0.35) & 79.93(0.07) \\
Moon     & \underline{54.21}(0.77)    & 81.37(0.27)      & 53.19(0.40)     & 69.69(0.31)    & 62.58(0.19)    & 84.75(0.01)    & 50.66(1.26)   & 61.88(0.82)  & 42.62(0.67) & 45.03(0.57) & 49.45(0.89) & 79.91(0.06)  \\
FedProx  & 53.98(0.56)    & 81.19(0.28)      & 54.83(0.83)     & 69.62(0.39)    & 63.54(0.27)    & 84.96(0.03)    & \underline{54.98}(0.91)   & 64.30(0.70)  & 43.66(0.91) & 46.15(0.91) & 49.72(0.86) & 79.93(0.06) \\ 
Scaffold & 53.92(1.60)    & 79.95(1.00)      & 56.74(0.68)     & \underline{70.98}(0.36)    & 65.10(1.25)    & 81.93(0.44)    & 48.68(2.23)   & 60.03(2.36)  & 30.42(1.50) & 36.46(0.96) & 45.07(2.04) & \underline{79.95}(0.01) \\ \midrule[0.3pt]
FedSage+ & 46.74(1.47)    & 80.75(0.45)      & \underline{57.14}(0.40)     & \textbf{73.22}(0.35)    & \underline{68.51}(0.02)    & \underline{86.06}(0.09)    & 54.05(1.08)   & 63.49(1.11)  & 42.82(1.04) & 46.46(1.68) & 54.43(0.34) & 73.35(1.32) \\
FedGL    & 41.77(1.12)    & 77.20(1.03)      & 41.11(1.29)     & 61.20(1.05)    & 61.97(0.63)    & 81.88(0.44)    & 48.21(4.40)   & 56.63(5.81)  & 30.52(3.74) & 33.14(2.84) & \underline{56.32}(1.23) & 70.27(2.24) \\
FedTAD  & 52.97(1.17)    & 81.40(0.41)      & 53.50(0.71)     & 68.45(0.75)    & 62.26(0.73)    & 84.64(0.16)    & 50.36(1.11)   & 61.26(0.96)  & 41.84(0.81) & 44.63(0.85) & 50.59(0.35) & 80.57(0.05) \\ \midrule[0.3pt]
Fed-PUB  & 47.13(1.24)    & 81.30(0.44)      & 52.15(0.88)     & 68.68(0.56)    & 66.03(2.01)    & 84.88(0.58)    & 49.81(5.24)   & 61.88(3.76)  & \underline{46.31}(0.72) & 43.30(1.15) & 47.87(0.69) & 79.36(2.05) \\
AdaFGL   & 52.15(0.98)    & \underline{82.67}(0.48)      & 54.82(0.51)     & 70.90(0.52)    & 62.79(0.38)    & 84.85(0.03)    & 51.40(0.98)   & 63.65(2.55)  & 43.53(1.58) & \underline{47.13}(1.48) & 49.12(0.40) & 79.92(0.08) \\
FedGTA   & 51.86(0.53)    & 81.06(0.36)      & 53.13(0.64)     & 68.43(0.38)    & 60.74(0.12)    & 84.22(0.01)    & 51.02(1.20)   & 60.69(0.91)  & 38.14(1.15) & 42.54(0.95) & 49.46(0.26) & 79.87(0.06) \\ \midrule[0.3pt]
FedSpray & 43.75(1.74)    & 72.13(2.17)      & 53.31(1.10)     & 67.71(1.06)    & 67.46(0.10)    & 83.26(0.05)    & 50.89(1.15)   & \underline{64.42}(1.00)  & 45.93(0.92) & \textbf{47.33}(1.28) &48.01(2.17) & 78.67(3.76) \\
FairFGL  & \textbf{57.37}(0.58)    & \textbf{82.85}(0.69)      & \textbf{58.23}(0.64)     & 70.87(0.58)    & \textbf{74.07}(0.38)    & \textbf{86.56}(0.19)    & \textbf{55.28}(1.01)   & \textbf{65.07}(1.19)  & \textbf{46.46}(1.09) & 46.46(0.98) & \textbf{58.69}(1.26) & \textbf{80.39}(0.25) \\ \midrule[0.3pt]
Improve  & \cellcolor{gray!25}{\textcolor{red}{$\Uparrow$} 5.83\%} & \cellcolor{gray!25}{\textcolor{red}{$\Uparrow$}} 0.22\% & \cellcolor{gray!25}{\textcolor{red}{$\Uparrow$} 1.91\%} & \cellcolor{gray!25}{NA} & \cellcolor{gray!25}{\textcolor{red}{$\Uparrow$} 8.07\%} & \cellcolor{gray!25}{\textcolor{red}{$\Uparrow$} 0.58\%} & \cellcolor{gray!25}{\textcolor{red}{$\Uparrow$} 2.28\%} & \cellcolor{gray!25}{\textcolor{red}{{$\Uparrow$}} 1.01\%} & \cellcolor{gray!25}{\textcolor{red}{$\Uparrow$} 1.15\%} & \cellcolor{gray!25}{NA} & \cellcolor{gray!25}{\textcolor{red}{$\Uparrow$} 4.21\%} & \cellcolor{gray!25}{\textcolor{red}{$\Uparrow$} 0.55\%} \\ \midrule[0.3pt]
\end{tabular}
}}
\end{table*}

\begin{table*}[t]
\caption{The notation of this Table aligns with Table~\ref{tab: Subgraph-FL Table 1}.}
\vspace{-0.1cm}
\label{tab: Subgraph-F1 Table 2}
\resizebox{\linewidth}{25mm}{
\setlength{\tabcolsep}{2.5mm}{
\begin{tabular}{c|ccc|ccc|ccc|ccc|}
\midrule[0.3pt]
Method   &  \multicolumn{3}{c|}{Cora}  &  \multicolumn{3}{c|}{CiteSeer}  &  \multicolumn{3}{c|}{PubMed}   & \multicolumn{3}{c|}{Tolokers} \\ \midrule[0.2pt]
Metrics  & Min-F1  & Hete-F1  & Hete-Min-F1  & Min-F1  & Hete-F1  & Hete-Min-F1  & Min-F1  & Hete-F1  & Hete-Min-F1  & Min-F1  & Hete-F1  & Hete-Min-F1   \\ \midrule[0.3pt]
FedAvg   & 69.75(2.41)    & 44.94(0.35)      & 31.39(0.80)     & 59.74(0.65)    & 41.77(0.61)    & 24.33(1.19)    & 63.84(0.15)   & 53.75(0.13)  & 45.34(2.07) & 47.85(0.53) & 49.56(0.32) & 59.05(0.62) \\
Moon     & 70.93(3.61)    & 45.37(1.16)      & 32.37(2.88)     & 59.88(0.77)    & 41.45(0.49)    & 24.66(1.37)    & 63.77(0.33)   & 53.61(0.12)  & 45.76(2.29) & 48.32(0.55) & 49.75(0.28) & 60.06(1.28)  \\
FedProx  & \underline{71.41}(1.55)    & 44.92(0.43)      & 31.71(2.27)     & 61.06(1.02)    & 42.77(0.98)    & 24.88(0.88)    & 64.64(0.20)   & 54.58(0.29)  & 48.08(2.78) & 47.86(0.32) & 49.57(0.21) & 59.81(1.00) \\ 
Scaffold & 68.57(0.38)    & \underline{45.50}(1.53)      & 30.48(6.53)     & 58.99(1.64)    & 44.61(1.02)    & 27.52(3.37)    & 63.86(0.96)   & 57.90(1.32)  & 49.20(1.98) & 43.66(0.13) & 40.99(0.11) & 51.60(0.70) \\ \midrule[0.3pt]
FedSage+ & 69.35(1.08)    & 37.55(1.17)     & 23.15(2.21)     & 56.76(1.76)    & \underline{46.43}(0.48)    & 26.20(1.52)    & \underline{72.42}(0.04)   & \underline{62.03}(0.03)  & \underline{58.47}(0.03) & 46.07(1.65) & 42.95(1.53) & 53.10(1.35) \\
FedGL    & 58.53(2.91)    & 37.39(1.84)      & 17.07(4.36)     & 54.23(1.25)    & 33.64(1.30)    & 25.99(2.09)    & 64.51(1.43)   & 52.66(0.65)  & 45.42(2.99) & 47.68(1.55) & \underline{50.80}(2.35) & \underline{61.65}(4.24) \\ 
FedTAD  & 68.69(2.93)    & 43.26(1.39)      & 28.74(2.41)     & 59.34(1.07)    & 42.02(0.62)    & 24.19(1.68)    & 63.33(0.83)   & 53.34(0.73)  & 44.37(1.54) & 48.41(0.59) & 49.66(0.59) & 59.92(0.78) \\ \midrule[0.3pt]
Fed-PUB  & 68.08(3.65)    & 40.07(0.98)      & 27.60(4.04)     & 60.13(0.90)    & 40.37(1.39)    & 23.79(1.96)    & 70.83(2.32)   & 58.87(1.67)  & 50.34(3.36) & 48.98(0.98) & 49.60(0.74) & 59.73(0.84) \\
AdaFGL   & 70.83(1.16)    & 43.52(1.32)      & \underline{32.54}(1.44)   & 60.55(1.04)    & 43.10(0.46)    & \underline{27.72}(2.06)    & 65.58(3.07)   & 54.91(1.65)  & 47.48(4.48) & 47.11(1.64) & 48.83(2.78) & 58.68(2.75) \\
FedGTA   & 70.72(1.23)    & 42.87(0.50)      & 30.86(2.21)     & 59.47(0.83)    & 41.40(0.83)    & 23.38(1.63)    & 61.69(0.12)   & 52.04(0.18)  & 41.95(0.96) & 47.80(0.44) & 49.64(0.25) & 60.06(0.43) \\ \midrule[0.3pt]
FedSpray & 64.26(2.74)    & 40.03(1.43)      & 27.56(5.84)     & \underline{61.20}(1.58)    & 43.35(1.68)    & 25.51(1.51)    & 67.55(3.85)   & 61.31(0.72)  & 55.86(6.12) & \underline{49.11}(0.58) & 47.62(0.61) & 57.22(1.22) \\
FairFGL  & \textbf{73.47}(1.19)    & \textbf{46.14}(0.75)      & \textbf{38.10}(2.04)     & \textbf{62.90}(0.98)    & \textbf{48.45}(1.23)    & \textbf{33.99}(2.48)    & \textbf{76.33}(0.62)   & \textbf{66.16}(0.52)  & \textbf{61.04}(1.36) & \textbf{49.68}(1.22) & \textbf{54.29}(0.31) & \textbf{66.33}(1.03) \\ \midrule[0.3pt]
Imporve & \cellcolor{gray!25}{\textcolor{red}{$\Uparrow$} 2.88 \%} & \cellcolor{gray!25}{\textcolor{red}{$\Uparrow$}} 1.41\% & \cellcolor{gray!25}{\textcolor{red}{$\Uparrow$} 17.09\%} & \cellcolor{gray!25}{\textcolor{red}{$\Uparrow$} 2.78\%} & \cellcolor{gray!25}{\textcolor{red}{$\Uparrow$} 4.35\%} & \cellcolor{gray!25}{\textcolor{red}{$\Uparrow$} 22.62\%} & \cellcolor{gray!25}{\textcolor{red}{$\Uparrow$} 5.4\%} & \cellcolor{gray!25}{\textcolor{red}{$\Uparrow$} 6.66\%}  & \cellcolor{gray!25}{\textcolor{red}{$\Uparrow$} 4.4\%} & \cellcolor{gray!25}{\textcolor{red}{$\Uparrow$} 1.16\%} & \cellcolor{gray!25}{\textcolor{red}{$\Uparrow$} 6.87\%} & \cellcolor{gray!25}{\textcolor{red}{$\Uparrow$}7.59\%} \\ \midrule[0.3pt]
\end{tabular}
}}
\end{table*}
Our experimental evaluation aims to address the following questions:
        \textbf{Q1}: Can FairFGL exhibit more comprehensive abilities in terms of offering fairer learning capabilities towards all nodes' categories? 
        \textbf{Q2}: How does each module contribute to FairFGL's success?
        \textbf{Q3}: Does the varying hyperparameter setting impact FairFGL's performance? 
        \textbf{Q4}: What is the efficiency of FairFGL in term of convergence speed? 
        
    \textbf{Experimental Environment}. Experiments are operated on the Ubuntu operating system with Intel(R) Xeon(R) Gold 5218R CPU @ 2.10GHz, 377GB memory, NVIDIA GeForce RTX 3090 with 24GB memory, and CUDA 12.0.

\subsection{Performance Comparison (Answer for Q1)}

The performance comparison results reported in Table~\ref{tab: Subgraph-FL Table 1} and Table~\ref{tab: Subgraph-F1 Table 2} exhibit a thorough evaluation of all methods in terms of their Fairness performances across all datasets.

\subsubsection{Overall Evaluation}
In Table~\ref{tab: Subgraph-FL Table 1}, we mainly report with two metrics including the Accuracy that is prevalently adopted as default metric indicating model's efficiency with node-level classification task. In general, FairFGL achieves competitive performance with baselines, but since FairFGL emphasizes on promoting Fairness Learning and most baselines excel at predicting labels for majority nodes, which composes large proportion of selected homogeneous graph datasets, the improvement made by FairFGL is subtle. 
In contrast, FairFGL demonstrates substantial gains in \textbf{Overall F1 score}, with improvements of 8.07\% on PubMed and 5.83\% on Cora. These results highlight a key divergence: while many baselines report robust accuracy, their F1 scores remain suboptimal. This observation supports our earlier discussion in Sec.~\ref{sec: Empirical Studies}, where we argued that relying solely on Accuracy can obscure a model’s poor performance on minority classes with limited data representation.
It is important to note that the Overall F1 metric includes all nodes—including those from the majority class—in its evaluation. This explains why some baselines, such as FedSage+ on CiteSeer, still achieve competitive F1 scores despite exhibiting bias. To better reflect performance disparities across demographic groups, Table~\ref{tab: Subgraph-F1 Table 2} highlights improvements specifically for disadvantaged node groups, particularly the \textbf{Hete-Min} category. These nodes are most severely affected by bias due to the disproportionate presence of \textbf{Homo-Maj} nodes in homophilic datasets. We focus our analysis on homophilic datasets for this reason: they present the most challenging setting for fairness-aware learning, as minority nodes are easily overwhelmed by structurally and semantically similar majority nodes. The following subsections provide detailed analysis across diverse baselines from different research domains.
\begin{table*}[t]  % 表格右浮动，占据半栏宽度
  \centering
  \caption{Ablation study performance (\%).}
  \label{tab: subgraph-fl ablation study}
    \resizebox{\linewidth}{!}{
\setlength{\tabcolsep}{2mm}{
\begin{tabular}{c|cccc|cccc}
\midrule[0.3pt]
Module              & \multicolumn{4}{c|}{CiteSeer}  & \multicolumn{4}{c}{Tolokers}    \\ \midrule[0.3pt] 
Metrics             & Overall F1 & Min-F1 & Hete-F1  & Hete-Min-F1 & Overall F1 & Min-F1 & Hete-F1  & Hete-Min-F1 \\ \midrule[0.3pt]
w/o History-Preserving  & 51.29(0.89) & 60.28(0.72) & 37.36(1.28) & 22.85(1.32) & 48.94(0.64) & 47.93(1.29) & 51.55(1.55) & 61.26(2.84) \\
w/o Gradient Modification   & 56.81(0.55) & 60.95(0.75) & 45.78(1.12) & 27.84(3.38) & 49.81(1.54) & 48.12(0.75) & 51.75(0.91) & 60.72(1.71) \\
w/o Clustering-based Aggregation & 57.61(0.55) & 62.72(0.79) & 46.31(0.75) & 30.08(1.32) & 48.03(0.29) & 47.85(0.25) & 48.81(0.21) & 58.62(0.77) \\
FairFGL             & 58.23(0.64) & 62.90(0.98) & 48.45(1.23) & 33.99(2.48) & 50.52(0.93) & 49.68(1.22) & 54.29(0.31) & 66.33(1.03)\\ \midrule[0.3pt]
\end{tabular}
}}
\end{table*}
\begin{figure*}[t]
  \includegraphics[width=\textwidth]{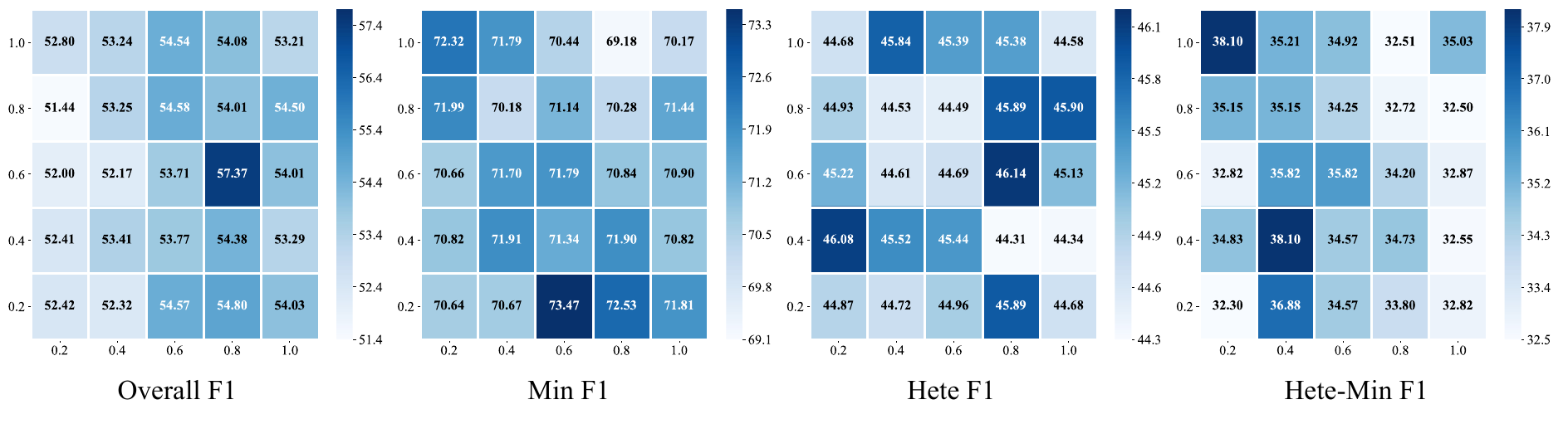}
  \captionsetup{font={small,stretch=1}}
  \caption{
  Hyperparameter Analysis on Cora Dataset.}
  
  \label{fig: hyperparameter}
  \vspace{-0.4cm}
\end{figure*}
\subsubsection{Compared with FL baselines}
Most FL methods report more competitive and resilient performance to Minority-class nodes compared with FGL baselines, which brings an interesting insight under the scenario of severe class-imbalanced distribution.
FL methods, originated from studies on Euclidean-space data type, are efficient in transferring class-wise knowledge in a collaborative training manner while FGL methods, due to their consideration for non-Euclidean space interactions among entities, suffer from the presumed Homophily Assumption.
However, this insight should not downplay the importance of FGL's development but rather reveal that most existing FGL methods, particularly for those aimed for training a generalizable global model, are not sufficient enough in handling Fairness challenge. Their designs are limited in acquiring the minority knowledge for its less consideration on heterophilous connectivity exhibited more often on nodes with minority class.
On the contrary, FairFGL manages to outperform all FL baselines for its consideration on both class and topology-wise perspectives.
The integration of the History-Preserving Module and Gradient Modification Module invites reliable global and cross-client minority knowledge learned from both perspectives. 
Specifically, FairFGL manages to reach 61.04\% precision on F1 score with improvements of 24.07\% compared to Scaffold in PubMed. In General, FL baselines present suboptimal performance on \textbf{Hete-Min} categories.  

\subsubsection{Compared with FGL baselines}
FGL baselines exhibit varied performance due to their distinct approaches to handling DH challenge. FedGL under-performs across all node categories, as it primarily addresses cross-client heterogeneity by compensating missing neighbors or labels, which overemphasize locally dominant majority-class nodes. In contrast, FedSage+ enhances performance by collaboratively training local generators to reconstruct missing neighbors, effectively restoring minority node connectivity. FedTAD transfers reliable class-wise knowledge derived from clients with high-homophilous local subgraphs, resulting in competitive outcomes. 

Personalized FGL approaches, such as FedPUB, AdaFGL, and FedSpray, exhibit improvements across most datasets for they manage to mitigate cross-client heterogeneity by adopting favoring training procedure locally with the learned global knowledge from the server-side aggregation. However, their methods pursue overall improvements that are reflect on their robust accuracy score under challenging class-imbalanced subgraph-FL settings, but lack of fine-grained exploration on node level, which leads to suboptimal performance on \textbf{Hete-Min} category. Similar argument can be made to FedGTA which performs the topology-aware aggregation and yield similar results with personalized FGL approach. In comparison, FairFGL manages to improve the F1 score on Hete-Min category with 22.62\% improvements on CiteSeer.

\subsubsection{Compared with FedSpray}
FedSpray aims to explore reliable structural proxies for sharing class-wise knowledge across clients. However, such a strategy does not directly address the biased local training processes affecting disadvantaged node groups. As a result, despite targeting class imbalance as its primary challenge, FedSpray does not achieve competitive performance compared with FairFGL.
In contrast, FairFGL incorporates both a Gradient Modification Module and a History-Preserving Module, explicitly designed to promote fair local training rather than merely transferring class-wise knowledge. Consequently, FairFGL delivers the most competitive performance across all node groups, particularly excelling on the severely underrepresented \textbf{Hete-Min} category.

\subsection{Ablation Studies (Answer for Q2)}

To answer Q2, we offer the ablation test showcasing how much impact does each key module contribute to the efficiency of FairFGL. 
In Table~\ref{tab: subgraph-fl ablation study}, we exclude each module separately and compare resulting performance with the full model on CiteSeer and Tolokers datasets. 
By excluding the History-Preserving Module, each local training solely involves the most recent updated global model at the inception of local training. Additionally, the knowledge distillation that provide consistent global knowledge on nodes with missing labels is excluded.  
Among all three modules, the History-Preserving Module exhibits most influence on FairFGL's overall performance in CiteSeer nearly across all metrics, which demonstrates the importance of regulating the local trained model from being over-fitted to the majority-dominated local data distribution when numbers of classes are increasing.
Notably, in CiteSeer, the History-Preserving Module contributes more on supplementing the topology-focused knowledge, which largely due to the knowledge distillation process. The injected minority knowledge via reconstructing the missing label information significantly enrich the local node profile, especially for \textbf{Hete-Min} nodes whose F1 score is improved with nearly 50\%. 
\begin{figure*}[t]
  \includegraphics[width=0.8\textwidth]{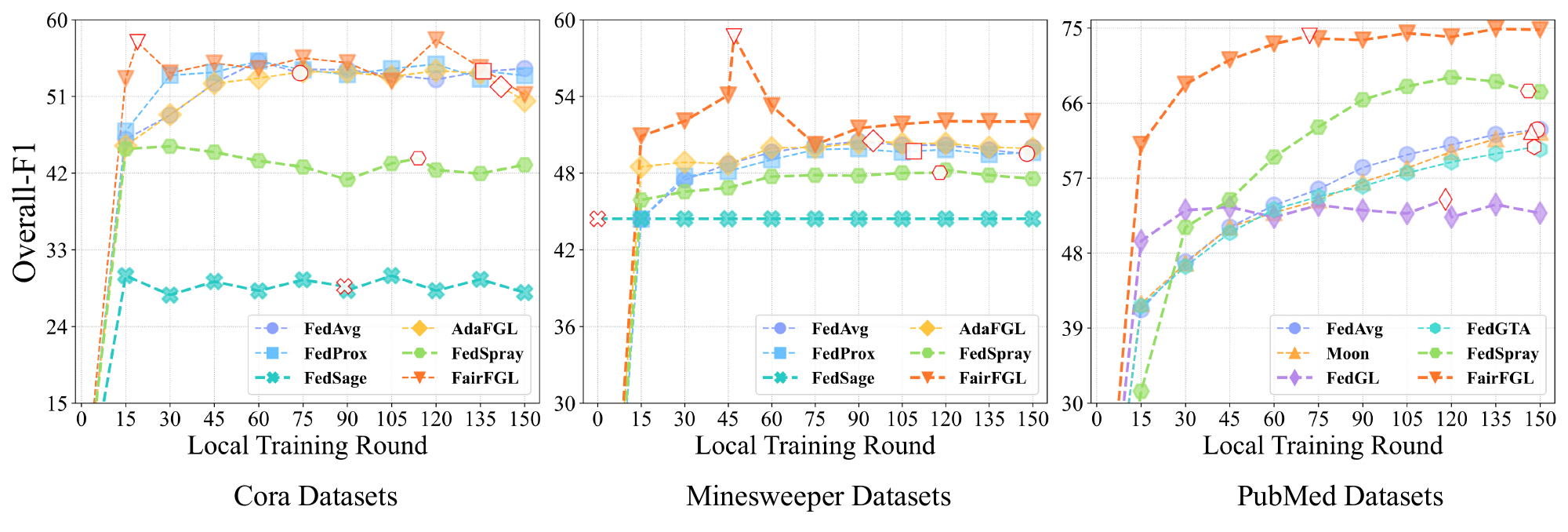}
  \centering
  \captionsetup{font={small,stretch=1}}
  \caption{
  Convergency Test on Cora, Minesweeper, and PubMed datasets.}
  
  \label{fig: convergence test}
  \vspace{-0.4cm}
\end{figure*}

Gradient Modification Module exhibits great improvements on FairFGL performance even if it is not as drastic as the other two modules, which is partially related to the regulatory term applied to prevent the Deviated Model from overwhelming the local training process. In other words, when History-Preserving Module already introduces key minority knowledge to the local training, the Gradient Modification Module functions as the supplementary role instead of the leading one. 

Clustering-based Aggregation is the most influential module with FairFGL's performance on Tolokers datasets. Differ from CiteSeer, Tolokers has limited labels but are densely connected, and therefore, the minority-class knowledge brought by the History-Preserving Module and Gradient Modification Module is limited. However, since topology becomes the main determinants to the parameter updates, the Clustering-based Aggregation Module manages to partition the clients to mitigate the topological heterogeneity, and consequently provides more reliable global knowledge for benefiting the other two modules. 

Overall, all three modules present significance to the improvement of \textbf{Here-Min} nodes the most, showcasing that FairFGL follows the intuitive principle of its project, which is to improve the model's equitable treatment towards nodes that are most biased due to the hostile local data distribution. For datasets that possess various class numbers, the History-Preserving Module brings more impact on models' performance, and for datasets that are densely connected, the Clustering-based Aggregation mechanism manages to mitigate the topological heterogeneity. 

\subsection{Hyperparameter Analysis (Answer for Q3)}
In FairFGL, there are two hyperparameters that respectively control the size of changed values of most influenced local parameters through client-side training (Top-$k$) and how much does each cluster generated at the server-side to contribute to yielding the optimal global update (Alpha). The details of the implementation of two hyperparameters have been thoroughly demonstrated in Sec.~\ref{sec: federated learning}. 
To evaluate how much of changes can different pairings of hyperparameters to the performance of FairFGL, we offer Fig.~\ref{fig: hyperparameter} on Cora datasets. Overall, FairFGL is not significantly impacted by the values of two hyperparameters, demonstrating both its robustness and its solid foundation on learning procedure that contributes to its success. 
Hyperparameter Top-$k$ influences the communication overhead and more importantly, further enhanced the privacy perservation of the silo data. Alpha functions mainly as the regulatory mechanism for selecting the best aggregation strategies.  
As the result propose that the overall variation of the FairFGL's performance is relatively small, and therefore, leave the manuever more freedom in operation. Notably, due to the wide scope of the evaluation metric, the optimal value of each does not share a single combination of two hyperparameters. This is due to the interwined nature of the definition of nodes' group that this paper propose. Nodes in datasets can not simply be exclusively partitioned into separate groups.  
However, the result does reflect a range of optimal combination as for choosing between 0.2 and 0.6 for Top-$k$, and 0.4 and 0.8 for Alpha, respectively.  For users that intend to maximize the privacy protection or communication efficiency, they can reduce the value of Top-$k$ and still acquire reasonably competitive performance, compared to results of baselines we offer in the Table~\ref{tab: Subgraph-FL Table 1}. 
Such analysis once emphasizes the robustness of FairFGL is acquired through its designed learning procedure. 

\subsection{Convergency Analysis (Answer for Q4)}
To evaluate the efficiency of FairFGL's learning process, we conduct a convergence analysis across three benchmark datasets, as illustrated in Fig.~\ref{fig: convergence test}. Specifically, we compare how quickly each method converges to this optimal point. For clarity, we annotate these optimal performance points in the figure using distinct hollow markers for each method.
Overall, FairFGL consistently reaches its optimal performance in fewer communication rounds compared to all baseline methods. This efficiency is primarily attributed to its communication mechanism, which reduces overhead by transmitting only the most influential parameters that encapsulate essential local knowledge. As outlined in our experimental settings, many baseline methods require over 100 rounds of federated training to converge. Consequently, we standardize the total number of training rounds to 150. While this uniform training budget results in some performance degradation for FairFGL beyond its convergence point, it highlights the framework’s ability to attain peak performance with significantly lower training cost.
Notably, in datasets such as PubMed, FairFGL maintains robust performance even as training continues, demonstrating its stability. For clarity and fairness in presentation, we selectively include combinations of representative baselines in the figure. This decision is made in part because certain methods, like FedSage+, incur substantially higher training time, making exhaustive comparisons computationally impractical. 

\section{Conclusion}
In this paper, we formally define the \textit{Fairness Challenge} in FGL, a critical yet largely overlooked aspect of data heterogeneity across clients, which refers to how biased training dynamics disproportionately impair underrepresented node groups.
To address this, we propose \textbf{FairFGL}, a fairness-centered FGL framework that explicitly regulates local training to prevent overfitting to dominant local classes. FairFGL leverages cross-client collaboration to transfer reliable class-wise knowledge, anchored in structurally confident topological patterns. Its improved communication and aggregation strategies balance efficiency and privacy preservation.
We advocate a promising direction to develop fairness-centered FGL frameworks tailored to heterophilous graphs, which encode rich and complex semantic relationships.
The experimental results demonstrate that FairFGL significantly outperforms SOTA baselines regarding fairness training with favorable training efficiency. 
%\section*{Acknowledgments}
%This should be a simple paragraph before the References to thank those individuals and institutions who have supported your work on this article.

\vfill

\newpage

\bibliographystyle{IEEEtran}
\bibliography{FairFGL}

\begin{thebibliography}{10}
\providecommand{\url}[1]{#1}
\csname url@samestyle\endcsname
\providecommand{\newblock}{\relax}
\providecommand{\bibinfo}[2]{#2}
\providecommand{\BIBentrySTDinterwordspacing}{\spaceskip=0pt\relax}
\providecommand{\BIBentryALTinterwordstretchfactor}{4}
\providecommand{\BIBentryALTinterwordspacing}{\spaceskip=\fontdimen2\font plus
\BIBentryALTinterwordstretchfactor\fontdimen3\font minus \fontdimen4\font\relax}
\providecommand{\BIBforeignlanguage}[2]{{%
\expandafter\ifx\csname l@#1\endcsname\relax
\typeout{** WARNING: IEEEtran.bst: No hyphenation pattern has been}%
\typeout{** loaded for the language `#1'. Using the pattern for}%
\typeout{** the default language instead.}%
\else
\language=\csname l@#1\endcsname
\fi
#2}}
\providecommand{\BIBdecl}{\relax}
\BIBdecl

\bibitem{infectious_Disease_Dynamics}
J.~L. Ke~Gu, Cao~Qian and Y.~Deng, ``Social privacy-preserving modeling based on graphical evolutionary game and infectious disease dissemination dynamics,'' \emph{{IEEE} Trans. Comput. Soc. Syst.}, vol.~11, no.~3, pp. 3882--3899, 2024.

\bibitem{xia2023app_gnn_rec1}
L.~Xia, C.~Huang, J.~Shi, and Y.~Xu, ``Graph-less collaborative filtering,'' in \emph{Proceedings of the ACM Web Conference, WWW}, 2023, pp. 17--27.

\bibitem{binwu_fina1}
B.~Wu, K.~Chao, and Y.~Li, ``Heterogeneous graph neural networks for fraud detection and explanation in supply chain finance,'' \emph{Inf. Syst.}, vol. 121, p. 102335, 2024.

\bibitem{hyun2023app_gnn_fina2}
W.~Hyun, J.~Lee, and B.~Suh, ``Anti-money laundering in cryptocurrency via multi-relational graph neural network,'' in \emph{Pacific-Asia Conference on Knowledge Discovery and Data Mining}.\hskip 1em plus 0.5em minus 0.4em\relax Springer, 2023, pp. 118--130.

\bibitem{Hu2021ahgae}
Y.~Hu, X.~Li, Y.~Wang, Y.~Wu, Y.~Zhao, C.~Yan, J.~Yin, and Y.~Gao, ``Adaptive hypergraph auto-encoder for relational data clustering,'' \emph{IEEE Transactions on Knowledge and Data Engineering}, 2021.

\bibitem{wu2019sgc}
F.~Wu, A.~Souza, T.~Zhang, C.~Fifty, T.~Yu, and K.~Weinberger, ``Simplifying graph convolutional networks,'' in \emph{International conference on machine learning, ICML}, 2019.

\bibitem{cai2021link_prediction2}
L.~Cai, J.~Li, J.~Wang, and S.~Ji, ``Line graph neural networks for link prediction,'' \emph{IEEE Transactions on Pattern Analysis and Machine Intelligence}, 2021.

\bibitem{tan2023link_prediction4}
Q.~Tan, X.~Zhang, N.~Liu, D.~Zha, L.~Li, R.~Chen, S.-H. Choi, and X.~Hu, ``Bring your own view: Graph neural networks for link prediction with personalized subgraph selection,'' in \emph{Proceedings of the Sixteenth ACM International Conference on Web Search and Data Mining, WSDM}, 2023, pp. 625--633.

\bibitem{zhang2019graph_classification1}
Z.~Zhang, J.~Bu, M.~Ester, J.~Zhang, C.~Yao, Z.~Yu, and C.~Wang, ``Hierarchical graph pooling with structure learning,'' \emph{arXiv preprint arXiv:1911.05954}, 2019.

\bibitem{yang2022graph_classification3}
M.~Yang, Y.~Shen, R.~Li, H.~Qi, Q.~Zhang, and B.~Yin, ``A new perspective on the effects of spectrum in graph neural networks,'' in \emph{International Conference on Machine Learning, ICML}.\hskip 1em plus 0.5em minus 0.4em\relax PMLR, 2022, pp. 25\,261--25\,279.

\bibitem{AdaFGL}
X.~Li, Z.~Wu, W.~Zhang, H.~Sun, R.-H. Li, and G.~Wang, ``Adafgl: A new paradigm for federated node classification with topology heterogeneity,'' in \emph{2024 IEEE 40th International Conference on Data Engineering (ICDE)}, 2024, pp. 2517--2530.

\bibitem{FedGTA}
\BIBentryALTinterwordspacing
X.~Li, Z.~Wu, W.~Zhang, Y.~Zhu, R.-H. Li, and G.~Wang, ``Fedgta: Topology-aware averaging for federated graph learning,'' \emph{VLDB Endowment}, vol.~17, no.~1, 2023. [Online]. Available: \url{https://doi.org/10.14778/3617838.3617842}
\BIBentrySTDinterwordspacing

\bibitem{Homophily_Assumption}
\BIBentryALTinterwordspacing
L.~Wu, H.~Lin, B.~Hu, C.~Tan, Z.~Gao, Z.~Liu, and S.~Z. Li, ``Beyond homophily and homogeneity assumption: Relation-based frequency adaptive graph neural networks,'' \emph{{IEEE} Trans. Neural Networks Learn. Syst.}, vol.~35, no.~6, pp. 8497--8509, 2024. [Online]. Available: \url{https://doi.org/10.1109/TNNLS.2022.3230417}
\BIBentrySTDinterwordspacing

\bibitem{Fedspray}
\BIBentryALTinterwordspacing
X.~Fu, Z.~Chen, B.~Zhang, C.~Chen, and J.~Li, ``Federated graph learning with structure proxy alignment,'' in \emph{Proceedings of the 30th ACM SIGKDD Conference on Knowledge Discovery and Data Mining}, ser. KDD '24.\hskip 1em plus 0.5em minus 0.4em\relax New York, NY, USA: Association for Computing Machinery, 2024, p. 827–838. [Online]. Available: \url{https://doi.org/10.1145/3637528.3671717}
\BIBentrySTDinterwordspacing

\bibitem{O-PFGL}
\BIBentryALTinterwordspacing
G.~Yan, X.~Li, L.~Xie, W.~Zhang, Q.~Shen, Y.~Fang, and Z.~Wu, ``Towards federated graph learning in one-shot communication,'' 2025. [Online]. Available: \url{https://arxiv.org/abs/2411.11304}
\BIBentrySTDinterwordspacing

\bibitem{FedSig}
\BIBentryALTinterwordspacing
B.~Bi, Z.~Zhang, P.~Qiao, Y.~Yuan, and G.~Wang, ``Fedsig: A federated graph augmentation for class-imbalanced node classification,'' in \emph{International Conference on Database Systems for Advanced Applications}.\hskip 1em plus 0.5em minus 0.4em\relax Springer-Verlag, 2024. [Online]. Available: \url{https://doi.org/10.1007/978-981-97-5552-3_32}
\BIBentrySTDinterwordspacing

\bibitem{2013firstgnn}
J.~Bruna, W.~Zaremba, A.~Szlam, and Y.~Lecun, ``Spectral networks and locally connected networks on graphs,'' \emph{Computer Science}, 2013.

\bibitem{kipf2016gcn}
T.~N. Kipf and M.~Welling, ``Semi-supervised classification with graph convolutional networks,'' in \emph{International Conference on Learning Representations, ICLR}, 2017.

\bibitem{atp}
\BIBentryALTinterwordspacing
X.~Li, J.~Ma, Z.~Wu, D.~Su, W.~Zhang, R.-H. Li, and G.~Wang, ``Rethinking node-wise propagation for large-scale graph learning,'' in \emph{Proceedings of the ACM Web Conference 2024}, ser. WWW '24.\hskip 1em plus 0.5em minus 0.4em\relax New York, NY, USA: Association for Computing Machinery, 2024, p. 560–569. [Online]. Available: \url{https://doi.org/10.1145/3589334.3645450}
\BIBentrySTDinterwordspacing

\bibitem{hamilton2017graphsage}
W.~Hamilton, Z.~Ying, and J.~Leskovec, ``Inductive representation learning on large graphs,'' \emph{Advances in neural information processing systems, NeurIPS}, 2017.

\bibitem{2019appnp}
J.~Klicpera, A.~Bojchevski, and S.~Günnemann, ``Predict then propagate: Graph neural networks meet personalized pagerank,'' in \emph{International Conference on Learning Representations, ICLR}, 2019.

\bibitem{wang2020gcnlpa}
H.~Wang and J.~Leskovec, ``Unifying graph convolutional neural networks and label propagation,'' \emph{arXiv preprint arXiv:2002.06755}, 2020.

\bibitem{chen2020gcnii}
M.~Chen, Z.~Wei, Z.~Huang, B.~Ding, and Y.~Li, ``Simple and deep graph convolutional networks,'' in \emph{International Conference on Machine Learning, ICML}, 2020.

\bibitem{mcmahan2017fedavg}
B.~McMahan, E.~Moore, D.~Ramage, S.~Hampson, and B.~A. y~Arcas, ``Communication-efficient learning of deep networks from decentralized data,'' \emph{Artificial intelligence and statistics}, pp. 1273--1282, 2017.

\bibitem{FGGP}
\BIBentryALTinterwordspacing
G.~Wan, W.~Huang, and M.~Ye, ``Federated graph learning under domain shift with generalizable prototypes,'' in \emph{Proceedings of the Thirty-Eighth AAAI Conference on Artificial Intelligence and Thirty-Sixth Conference on Innovative Applications of Artificial Intelligence and Fourteenth Symposium on Educational Advances in Artificial Intelligence}, ser. AAAI'24/IAAI'24/EAAI'24.\hskip 1em plus 0.5em minus 0.4em\relax AAAI Press, 2024. [Online]. Available: \url{https://doi.org/10.1609/aaai.v38i14.29468}
\BIBentrySTDinterwordspacing

\bibitem{FGSSL}
W.~Huang, G.~Wan, M.~Ye, and B.~Du, ``Federated graph semantic and structural learning,'' \emph{IJCAI}, 2023.

\bibitem{baek2022fedpub}
J.~Baek, W.~Jeong, J.~Jin, J.~Yoon, and S.~J. Hwang, ``Personalized subgraph federated learning,'' \emph{arXiv preprint arXiv:2206.10206}, 2023.

\bibitem{feddep}
\BIBentryALTinterwordspacing
K.~Zhang, L.~Sun, B.~Ding, S.~M. Yiu, and C.~Yang, ``Deep efficient private neighbor generation for subgraph federated learning,'' 2024. [Online]. Available: \url{https://arxiv.org/abs/2401.04336}
\BIBentrySTDinterwordspacing

\bibitem{fedtad}
\BIBentryALTinterwordspacing
Y.~Zhu, X.~Li, Z.~Wu, D.~Wu, M.~Hu, and R.-H. Li, ``Fedtad: Topology-aware data-free knowledge distillation for subgraph federated learning,'' in \emph{Proceedings of the Thirty-Third International Joint Conference on Artificial Intelligence, {IJCAI-24}}, K.~Larson, Ed.\hskip 1em plus 0.5em minus 0.4em\relax International Joint Conferences on Artificial Intelligence Organization, 8 2024, pp. 5716--5724, main Track. [Online]. Available: \url{https://doi.org/10.24963/ijcai.2024/632}
\BIBentrySTDinterwordspacing

\bibitem{fairness_FL_Survey}
Y.~Shi, H.~Yu, and C.~Leung, ``Towards fairness-aware federated learning,'' \emph{IEEE Transactions on Neural Networks and Learning Systems}, vol.~35, no.~9, pp. 11\,922--11\,938, 2024.

\bibitem{BalanceFL}
X.~Shuai, Y.~Shen, S.~Jiang, Z.~Zhao, Z.~Yan, and G.~Xing, ``Balancefl: Addressing class imbalance in long-tail federated learning,'' in \emph{2022 21st ACM/IEEE International Conference on Information Processing in Sensor Networks (IPSN)}, 2022, pp. 271--284.

\bibitem{FedeAMC}
Y.~Wang, G.~Gui, H.~Gacanin, B.~Adebisi, H.~Sari, and F.~Adachi, ``Federated learning for automatic modulation classification under class imbalance and varying noise condition,'' \emph{IEEE Transactions on Cognitive Communications and Networking}, vol.~8, no.~1, pp. 86--96, 2022.

\bibitem{fairness_factorization}
S.~Liu, Y.~Ge, S.~Xu, Y.~Zhang, and A.~Marian, ``Fairness-aware federated matrix factorization,'' in \emph{Association for Computing Machinery}, New York, NY, USA, 2022.

\bibitem{FAPL}
X.-X. Wei and H.~Huang, ``Balanced federated semisupervised learning with fairness-aware pseudo-labeling,'' \emph{IEEE Transactions on Neural Networks and Learning Systems}, vol.~35, no.~7, pp. 9395--9407, 2024.

\bibitem{flfaultD}
S.~Sun, J.~Qian, X.~Zhang, and Z.~Song, ``Class-imbalance and client-imbalance federated learning for fault diagnosis,'' in \emph{2024 IEEE 13th Data Driven Control and Learning Systems Conference (DDCLS)}, 2024, pp. 860--865.

\bibitem{CUCB}
M.~Yang, X.~Wang, H.~Zhu, H.~Wang, and H.~Qian, ``Federated learning with class imbalance reduction,'' in \emph{2021 29th European Signal Processing Conference (EUSIPCO)}, 2021, pp. 2174--2178.

\bibitem{Fed-Focal_Loss}
\BIBentryALTinterwordspacing
D.~Sarkar, A.~Narang, and S.~Rai, ``Fed-focal loss for imbalanced data classification in federated learning,'' 2020. [Online]. Available: \url{https://arxiv.org/abs/2011.06283}
\BIBentrySTDinterwordspacing

\bibitem{FRAug}
H.~Chen, A.~Frikha, D.~Krompass, J.~Gu, and V.~Tresp, ``Fraug: Tackling federated learning with non-iid features via representation augmentation,'' in \emph{Proceedings of the IEEE/CVF International Conference on Computer Vision (ICCV)}, October 2023, pp. 4849--4859.

\bibitem{Astraea}
M.~Duan, D.~Liu, X.~Chen, R.~Liu, Y.~Tan, and L.~Liang, ``Self-balancing federated learning with global imbalanced data in mobile systems,'' \emph{IEEE Transactions on Parallel and Distributed Systems}, vol.~32, no.~1, pp. 59--71, 2021.

\bibitem{GraphSmote}
T.~Zhao, X.~Zhang, and S.~Wang, ``Graphsmote: Imbalanced node classification on graphs with graph neural networks,'' in \emph{Association for Computing Machinery}, 2021.

\bibitem{FedLog}
\BIBentryALTinterwordspacing
S.~Kim, Y.~Lee, C.~Yang, Y.~Oh, N.~Lee, S.~Yun, J.~Lee, S.~Kim, and C.~Park, ``Subgraph federated learning for local generalization,'' in \emph{FedKDD: International Joint Workshop on Federated Learning for Data Mining and Graph Analytics}, 2024. [Online]. Available: \url{https://openreview.net/forum?id=dCrqPTrF7y}
\BIBentrySTDinterwordspacing

\bibitem{FGPL}
X.~Kong, H.~Yuan, G.~Shen, H.~Zhou, W.~Liu, and Y.~Yang, ``Mitigating data imbalance and generating better prototypes in heterogeneous federated graph learning,'' \emph{Knowledge-Based Systems}, vol. 296, p. 111876, 2024.

\bibitem{Pan2023TowardsFG}
C.~Pan, J.~Xu, Y.~Yu, Z.~Yang, Q.~Wu, C.~Wang, L.~Chen, and Y.~Yang, ``Towards fair graph federated learning via incentive mechanisms,'' in \emph{AAAI}, vol.~38, no.~13, 2024, pp. 14\,499--14\,507.

\bibitem{NoPrejudice}
\BIBentryALTinterwordspacing
N.~Agrawal, A.~K. Sirohi, S.~Kumar, and J.~, ``No prejudice! fair federated graph neural networks for personalized recommendation,'' \emph{Proceedings of the AAAI Conference on Artificial Intelligence}, vol.~38, no.~10, pp. 10\,775--10\,783, Mar. 2024. [Online]. Available: \url{https://ojs.aaai.org/index.php/AAAI/article/view/28950}
\BIBentrySTDinterwordspacing

\bibitem{PFedNC}
Q.~Mao, X.~Lin, G.~Li, L.~Chen, Y.~Liu, Y.~Qi, and J.~Li, ``Ensuring minority group rights in social iot with fairness-aware federated graph node classification,'' in \emph{2023 IEEE Intl Conf on Parallel \& Distributed Processing with Applications, Big Data \& Cloud Computing, Sustainable Computing \& Communications, Social Computing \& Networking (ISPA/BDCloud/SocialCom/SustainCom)}, 2023, pp. 125--130.

\bibitem{Graph_pruning}
S.~Han, J.~Pool, J.~Tran, and W.~J. Dally, ``Learning both weights and connections for efficient neural networks,'' in \emph{Proceedings of the 28th International Conference on Neural Information Processing Systems - Volume 1}, ser. NIPS'15.\hskip 1em plus 0.5em minus 0.4em\relax Cambridge, MA, USA: MIT Press, 2015, p. 1135–1143.

\bibitem{Predict_Parameters}
M.~Denil, B.~Shakibi, L.~Dinh, M.~Ranzato, and N.~de~Freitas, ``Predicting parameters in deep learning,'' in \emph{Proceedings of the 26th International Conference on Neural Information Processing Systems - Volume 2}, ser. NIPS'13.\hskip 1em plus 0.5em minus 0.4em\relax Red Hook, NY, USA: Curran Associates Inc., 2013, p. 2148–2156.

\bibitem{openfgl}
\BIBentryALTinterwordspacing
X.~Li, Y.~Zhu, B.~Pang, G.~Yan, Y.~Yan, Z.~Li, Z.~Wu, W.~Zhang, R.-H. Li, and G.~Wang, ``Openfgl: A comprehensive benchmark for federated graph learning,'' 2025. [Online]. Available: \url{https://arxiv.org/abs/2408.16288}
\BIBentrySTDinterwordspacing

\bibitem{zhang2021fedsage}
K.~Zhang, C.~Yang, X.~Li, L.~Sun, and S.~M. Yiu, ``Subgraph federated learning with missing neighbor generation,'' \emph{Advances in Neural Information Processing Systems, NeurIPS}, 2021.

\bibitem{chen2021fedgl}
C.~Chen, W.~Hu, Z.~Xu, and Z.~Zheng, ``Fedgl: federated graph learning framework with global self-supervision,'' \emph{arXiv preprint arXiv:2105.03170}, 2021.

\bibitem{karimireddy2020scaffold}
S.~P. Karimireddy, S.~Kale, M.~Mohri, S.~Reddi, S.~Stich, and A.~T. Suresh, ``Scaffold: Stochastic controlled averaging for federated learning,'' in \emph{International Conference on Machine Learning, ICML}.\hskip 1em plus 0.5em minus 0.4em\relax PMLR, 2020, pp. 5132--5143.

\bibitem{li2020fedporx}
T.~Li, A.~K. Sahu, M.~Zaheer, M.~Sanjabi, A.~Talwalkar, and V.~Smith, ``Federated optimization in heterogeneous networks,'' \emph{Proceedings of Machine Learning and Systems, MLSys}, vol.~2, pp. 429--450, 2020.

\bibitem{li2021moon}
Q.~Li, B.~He, and D.~Song, ``Model-contrastive federated learning,'' in \emph{Proceedings of the IEEE/CVF Conference on Computer Vision and Pattern Recognition, CVPR}, 2021, pp. 10\,713--10\,722.

\bibitem{akiba2019optuna}
T.~Akiba, S.~Sano, T.~Yanase, T.~Ohta, and M.~Koyama, ``Optuna: A next-generation hyperparameter optimization framework,'' in \emph{Proceedings of the 25th ACM SIGKDD international conference on knowledge discovery \& data mining, KDD}, 2019, pp. 2623--2631.

\end{thebibliography}

\end{document}